\documentclass[11pt]{article}

\usepackage{acl}
\usepackage{amsmath}
\usepackage{amsfonts}
\usepackage{amsthm}

\usepackage{times}
\usepackage{latexsym}
\usepackage{subcaption}
\usepackage{enumitem}
\usepackage{booktabs}
\usepackage{multirow}

\newtheorem{theorem}{Theorem}[section]      % Theorem numbering by section
\newtheorem{proposition}[theorem]{Proposition}

\newtheorem{assumption}[theorem]{Assumption}

\usepackage{times}
\usepackage{latexsym}
\usepackage{ulem}
\usepackage{xcolor}
\usepackage{soul}
\setulcolor{blue}
\usepackage[T1]{fontenc}
\usepackage[utf8]{inputenc}
\usepackage{ulem} 
\usepackage{microtype}
\usepackage{multirow}
\usepackage{makecell}
\usepackage{booktabs}
\usepackage{graphicx}
\usepackage{amssymb}
\usepackage{url}
\usepackage{hyperref}
\usepackage{caption}
\usepackage[ruled,vlined,linesnumbered]{algorithm2e}
\usepackage{xcolor}

\usepackage{rotating}

\usepackage{inconsolata}

\usepackage{graphicx}
\newcommand{\boldhdr}[1]{\vspace{0.05cm}\noindent \textbf{#1.}}

\title{Why LoRA Fails to Forget: Regularized Low-Rank Adaptation Against Backdoors in Language Models}
\author{Hoang-Chau Luong \ \ \ \  Lingwei Chen \\
  Golisano College of Computing and Information Sciences \\
  Rochester Institute of Technology \\
  Rochester, NY, United States \\
  \texttt{cl6300@rit.edu, lwcics@rit.edu} \\\\}

\begin{document}
\maketitle

\begin{abstract}
    Low-Rank Adaptation (LoRA) is widely used for parameter-efficient fine-tuning of large language models, but it is notably ineffective at removing backdoor behaviors from poisoned pretrained models when fine-tuning on clean dataset. Contrary to the common belief that this weakness is caused primarily by low rank, we show that LoRA's vulnerability is fundamentally spectral. Our analysis identifies two key factors: LoRA updates (i) possess insufficient spectral strength, with singular values far below those of pretrained weights, and (ii) exhibit unfavorable spectral alignment, weakly matching clean-task directions while retaining overlap with trigger-sensitive subspaces. We further establish a critical scaling threshold beyond which LoRA can theoretically suppress trigger-induced activations, and we show empirically that standard LoRA rarely reaches this regime. We introduce Regularized Low-Rank Adaptation (RoRA), which improves forgetting by increasing spectral strength and correcting alignment through clean-strengthened regularization, trigger-insensitive constraints, and post-training spectral rescaling. Experiments across multiple NLP benchmarks and attack settings show that RoRA substantially reduces attack success rates while maintaining clean accuracy.
\end{abstract}

\section{Introduction}

Large language models (LLMs) have achieved remarkable success across a wide range of natural language processing (NLP) tasks \citep{brown2020language, raffel2020exploring, devlin2019bert}. 
However, the rapid growth in model size has rendered full fine-tuning increasingly expensive~\citep{ding2023parameter}, motivating the development of parameter-efficient fine-tuning (PEFT) methods that adapt pretrained models with substantially reduced both computational and memory overhead \citep{houlsby2019parameter, lester2021power, hu2022lora}. Among these approaches, Low-Rank Adaptation (LoRA) \citep{hu2022lora} has emerged as a de facto standard, reparameterizing the weight updates into low-rank matrices that enable efficient adaptation while preserving strong downstream performance.

Despite their practical advantages, recent studies have exposed a critical vulnerability of LoRA to backdoor attacks. Empirical evidence shows that models fine-tuned with LoRA often retain trigger-induced behaviors even after adaptation to downstream tasks~\citep{gu2023gradient, zhao2024defending, zhao2024unlearning}. 
In contrast, full fine-tuning (FFT) can partially mitigate backdoor effects by directly overwriting corrupted parameters. Most existing defenses attribute LoRA's limited robustness primarily to its low-rank constraint, arguing that reduced rank inherently restricts representational capacity compared to full-rank fine-tuning~\cite{sun2023defending, zhao2024defending, zhao2024unlearning, zhao2025breakingpeftlimitationsleveraging}. 
This perspective has motivated a range of external remedies, including trigger detectors, poisoned sample detections, auxiliary clean models for knowledge distillation, and multi-stage fine-tuning pipelines. 
However, these approaches offer limited insight into why LoRA fails to forget backdoor triggers and provide little principled guidance for improving LoRA itself. 
Consequently, the intrinsic forgetting dynamics of LoRA remain poorly understood.

In this work, we revisit the forgetting behavior of LoRA from a spectral perspective and challenge the prevailing assumption that low rank is the primary bottleneck. We argue instead that LoRA’s vulnerability arises from two interacting factors: \textit{insufficient spectral strength} of its updates and \textit{unfavorable alignment} of their representations with downstream and trigger-sensitive directions. We first establish, theoretically, the existence of a critical scaling threshold on the LoRA updates when added to the pretrained weights. Beyond this threshold, the low-rank component can dominate trigger-induced activations and suppress backdoor behavior. However, our empirical analysis shows that standard LoRA updates typically have maximum singular values far smaller than those of the pretrained weights, making the required threshold difficult to attain in practice. In addition, LoRA updates tend to align weakly with clean-task directions while remaining partial alignment with spectral subspaces of the pretrained model, which may overlap with trigger-sensitive directions and further hinder effective forgetting.
% : \textcolor{red}{pretrained weights already encode strong downstream information, reducing the incentive for LoRA to learn clean features, while the remaining nonzero alignment can amplify trigger responses rather than suppress them.}

Motivated by these observations, we propose Regularized Low-Rank Adaptation (RoRA), a principled enhancement of LoRA that improves backdoor robustness through three complementary mechanisms:
(i) \textit{clean-strengthened regularization}, which applies dropout to the pretrained weights to encourage the LoRA updates to learn stronger clean-task directions;
(ii) \textit{trigger-insensitive regularization}, which enforces orthogonality between the LoRA updates and pretrained spectral subspaces, preventing the relearning of poisoned directions; and (iii) \textit{post-training spectral rescaling}, which adaptively increases the effective spectral strength of the LoRA updates to compete with the pretrained model's leading spectral modes.

Our contributions are summarized as follows:
\begin{enumerate}[leftmargin=12pt]
    \vspace{-5pt}
    \item We revisit LoRA's forgetting behavior through a spectral lens and show that LoRA's vulnerability arises not primarily from low rank, but from insufficient spectral strength and unfavorable alignment of its updates.

    \vspace{-7pt}
    \item We develop a theoretical analysis of backdoor forgetting under LoRA, proving the existence of a critical scaling threshold at which the LoRA updates dominate trigger-induced activations.

    \vspace{-7pt}
    \item We propose RoRA, a principled enhancement of LoRA that improves backdoor robustness via complementary regularizations and post-training spectral rescaling, achieving substantially lower attack success rates while preserving high clean accuracy, and can also be applied to other LoRA variants for robustness improvement.
\end{enumerate}

\section{Related works}

\textbf{Backdoor attacks}, initially presented in computer vision~\citep{hu2022badhash}, have recently garnered interest in NLP~\citep{dong2021should, li-etal-2021-bfclass-backdoor, zhou2024backdoor} due to their implications for language model security~\citep{dong2021towards, formento2023using, minh2022textual}. In a typical backdoor attack, attackers insert rare words or sentences into input samples as triggers and modify their labels to induce malicious behavior~\citep{qi2021hidden, chen2021mitigating}. BadNet~\citep{gu2017badnets} inserts rare character sequences (e.g., ``mn''), and \citet{chen2021mitigating} extend this idea by employing infrequent words as triggers. InSent~\citep{dai2019backdoor} instead uses fixed sentences to trigger targeted predictions, while \citet{qi2021hidden} exploit syntactic structures as attack triggers to enhance stealthiness. \citet{zhao2024weak} enhance backdoor attacks on large-scale models fine-tuned with PEFT by applying knowledge distillation from a small teacher model to a large-scale student model.

\boldhdr{Backdoor defenses}
Backdoor defenses in NLP primarily aim to disrupt trigger activation while preserving clean performance. ONION~\cite{qi-etal-2021-onion} detects suspicious tokens by measuring their influence on perplexity and filters those that introduce abnormal changes. Back-Translation~\cite{qi2021hidden} reduces trigger effectiveness by paraphrasing inputs through translation. SCPD~\cite{qi2021hidden} restructures input sentences using predefined syntactic templates to diminish trigger impact. A complementary line of work develops defenses under PEFT. \citet{zhao2024defending} randomize labels and fine-tune poisoned models with PEFT modules, identifying poisoned samples via model confidence; however, the method requires multiple fine-tuning rounds, is computationally costly, and does not improve the model itself. \citet{zhao2024unlearning} introduce an unlearning framework that removes backdoor behavior via knowledge distillation, but it relies on constructing a clean teacher model, which is often impractical.

\vspace{-2pt}
\section{Preliminaries}
\vspace{-3pt}

\boldhdr{Threat model and attack capability} 
We focus on weight-poisoning backdoor attacks, in which an adversary embeds trigger-label associations into the pretrained parameters so that inputs containing a specific trigger are mapped to an adversary-chosen target label \citep{li-etal-2021-bfclass-backdoor, du2022ppt, xu-etal-2022-exploring, sun2023defending, zhao2024defending}. We assume that a pretrained language model is poisoned during pretraining, and study whether downstream fine-tuning strategies can overwrite such poisoning. The adversary has partial knowledge of the downstream task but no access to the target dataset or its training instances. Thus, the poisoned model is constructed using a proxy dataset with a similar label distribution. This reflects realistic scenarios where task information is more accessible, while downstream data remain private.

\boldhdr{Problem definition}
Without loss of generability, let $\mathbf{W}_{\text{pre}}$ denote the weights of a poisoned language model, and let $(\mathbf{x}, y) \in \mathcal{D}^{\text{train}}_{\text{clean}}$ and $(\mathbf{x}, y) \in \mathcal{D}^{\text{test}}_{\text{clean}}$ be the clean training and test datasets, respectively. Given a target input $\mathbf{x}^{\rm{trig}}$ containing the pre-defined trigger, the attacker aims to force the poisoned model to misclassify $\mathbf{x}^{\rm{trig}}$ as the target label $y_{\rm{bd}}$. Our defense goal is to ensure that, after fine-tuning $\mathbf{W}_{\text{pre}}$ on $\mathcal{D}^{\text{train}}_{\text{clean}}$ using LoRA, the model can predict the correct label for $\mathbf{x}^{\rm{trig}}$. In other words, we seek to enhance downstream fine-tuning to effectively overwrite backdoor behaviors embedded in the pretrained weights.

\vspace{-3pt}
\section{Backdoor Forgetting in LoRA}
\label{sec:main_analysis}
\vspace{-3pt}

\subsection{LoRA versus Full Fine-tuning}
\vspace{-3pt}

\boldhdr{Full fine-tuning forgets more than LoRA}
Full fine-tuning (FFT) updates all parameters of a pretrained model, whereas LoRA introduces only a small number of trainable low-rank parameters. Since FFT directly overwrites the poisoned weights $\mathbf{W}_{\text{pre}}$, it has substantially greater capacity to erase backdoor representations and suppress backdoor behavior when trained on clean data~\citep{sun2023defending,zhao2024unlearning}. In contrast, LoRA often preserves trigger-induced behavior even after adaptation to downstream task~\citep{gu2023gradient,zhao2024defending,zhao2024unlearning}. The stronger forgetting behavior of FFT is closely related to catastrophic forgetting~\citep{mccloskey1989catastrophic}: optimization toward the downstream objective may overwrite associations learned earlier, including those responsible for trigger-target mappings.

% \boldhdr{LoRA forgets less: beyond the rank constraint}
LoRA reduces the memory and compute cost by restricting adaptation to a low-rank update.
For a linear layer $\mathbf{W}_{\text{pre}}\in\mathbb{R}^{d\times d}$, LoRA produces
\begin{equation}
\label{eq:lora}
    \mathbf{W}' = \mathbf{W}_{\text{pre}} + s \Delta \mathbf{W},
\end{equation}
where $s = \alpha/r$ is a scaling factor with $\alpha$ is a hyperparameter and $\Delta \mathbf{W}=\mathbf{B}\mathbf{A}$ with $\mathbf{A}\in\mathbb{R}^{r\times d}$, $\mathbf{B}\in\mathbb{R}^{d\times r}$, and $r\ll d$.
This reduces trainable parameters from $d^2$ to $2dr$. A common assumption for LoRA's weaker backdoor forgetting is that the low-rank constraint reduces expressive capacity~\citep{zhao2024defending, zhao2024unlearning}.
However, low rank alone does not well explain when or why LoRA fails. \citet{shuttleworth2025lora} suggests that LoRA can still forget pretrained knowledge under certain conditions in continual-learning settings. This motivates a mechanistic question central to our work:
\textit{under what conditions can a low-rank update suppress trigger-induced logits?}

\boldhdr{Backdoor suppression via LoRA}
To reason about backdoor suppression under LoRA, we rewrite Eq.~\eqref{eq:lora}, which yields
\vspace{-2pt}
\begin{equation}
\label{eq:first_decom}
    \mathbf{W}' = \mathbf{W}_{\mathrm{clean}} + \mathbf{W}_{\mathrm{pois}} + s \Delta\mathbf{W},
\vspace{-2pt}
\end{equation}
where $\mathbf{W}_{\mathrm{clean}}$ and $\mathbf{W}_{\mathrm{pois}}$ denote the clean-task and backdoor-related components of the pretrained weights, respectively.
This decomposition suggests two conceptually distinct mechanisms:
\begin{itemize}[leftmargin=10pt]
    \vspace{-6pt}
    \item \textit{Direct cancellation:} $\Delta\mathbf{W}$ explicitly suppresses $\mathbf{W}_{\mathrm{pois}}$, neutralizing trigger-induced activations.
    This requires detecting backdoor-relevant directions, which is typically difficult without additional modules or assumptions.

    \vspace{-8pt}
    \item \textit{Task dominance:} $\Delta\mathbf{W}$ amplifies clean representations so that, at inference time, clean logits dominate those induced by the trigger.

    \vspace{-6pt}
\end{itemize}
In this work, we focus exclusively on task dominance: we seek conditions under which LoRA can suppress backdoor behavior by amplifying clean-task margins, without requiring trigger detection.

\vspace{-3pt}
\subsection{Spectral Analysis}
\vspace{-4pt}

\boldhdr{A spectral lens}
We apply singular value decomposition (SVD) to rewrite the pretrained weights and the LoRA update as $\mathbf{W}_{\mathrm{pre}} = \sum_i \sigma^{i}_{\mathrm{pre}}\, \mathbf{u}^{i}_{\mathrm{pre}}(\mathbf{v}^{i}_{\mathrm{pre}})^{\top}$, and $\Delta\mathbf{W} = \sum_i \sigma^{i}_{\Delta}\, \mathbf{u}^{i}_{\Delta}(\mathbf{v}^{i}_{\Delta})^{\top}$. This decomposition reveals how the low-rank update interacts with the dominant spectral modes of the pretrained model. Building on this perspective, we analyze LoRA's capacity to mitigate backdoor attacks through a task-dominance mechanism governed by two factors that are spectral strength and spectral alignment. Our analysis shows that LoRA suppresses trigger-induced behavior when both following conditions are satisfied: (1) the update has sufficient spectral strength to influence the decision boundary, and (2) its dominant directions align with clean-task structure while remaining weakly aligned with backdoor-sensitive directions.

\begin{assumption}
\label{assump:smax-alignment}
We consider an arbitrary clean input-label pair $(\mathbf{x},y)$ and its triggered counterpart $\mathbf{x}^{\mathrm{trig}}$. 
We assume both inputs are $\ell_2$-normalized:
$
    \|\mathbf{x}\|_2 = \|\mathbf{x}^{\mathrm{trig}}\|_2 = 1.
$
Let $\mathbf{W}_{\mathrm{pre}}\in\mathbb{R}^{C\times d}$ be a poisoned frozen linear classifier
and $\Delta\mathbf{W}\in\mathbb{R}^{C\times d}$ a LoRA update, where $C$ denotes the number of classes. For $s>0$, define
\vspace{-2pt}
\begin{align}
    z_s(\mathbf{x}) &:= (\mathbf{W}_{\mathrm{pre}} + s\,\Delta\mathbf{W})\mathbf{x},
    \\
    M_s(\mathbf{x}) &:= z_s(\mathbf{x})_y - z_s(\mathbf{x})_{y_{\mathrm{bd}}}
\vspace{-2pt}
\end{align}
with spectral norms (maximum singular values) written as
$\sigma_{\mathrm{pre}}\!:=\!\|\mathbf{W}_{\mathrm{pre}}\|_2\!=\!\sigma_{\max}(\mathbf{W}_{\mathrm{pre}}),
\
\sigma_{\Delta}\!:=\!\|\Delta\mathbf{W}\|_2\!=\!\sigma_{\max}(\Delta\mathbf{W}).
$ 
For logit coordinates, let $\mathbf{e}_k\in\mathbb{R}^C$ denote the canonical selector, so that 
\(\langle \mathbf{e}_k,\mathbf{z}\rangle = z_k\) for any logit vector $\mathbf{z}$.
The margin direction between the clean label $y$ and the backdoor target $y_{\mathrm{bd}}$ is defined as
$
\mathbf{c} := \mathbf{e}_y - \mathbf{e}_{y_{\mathrm{bd}}}.
$
We assume

\vspace{0.2cm}\noindent\textnormal{(A1)} \textit{Effective backdoor at initialization.}\label{assump:A1}
There exists $\rho_{\mathrm{bd}}\in(0,1]$ such that
\begin{equation}
\label{eq:assumption1}
    \langle \mathbf{c},\,\mathbf{W}_{\mathrm{pre}}\mathbf{x}^{\mathrm{trig}}\rangle
    \le -\,\rho_{\mathrm{bd}}\,\|\mathbf{c}\|_2\,\sigma_{\mathrm{pre}} < 0.
\end{equation}
\noindent\textnormal{(A2)} \textit{LoRA's clean-trigger margin.}
There exist $\rho_{\mathrm{cl}}\in(0,1]$ and $\rho_{\mathrm{tr}}\in[0,1)$ such that
\begin{align}
    &\langle \mathbf{c},\,\Delta\mathbf{W}\mathbf{x}\rangle
    \; \ge \; \rho_{\mathrm{cl}}\,\|\mathbf{c}\|_2\,\sigma_{\Delta}, \\
    &\bigl|\langle \mathbf{c},\,\Delta\mathbf{W}(\mathbf{x}^{\mathrm{trig}}-\mathbf{x})\rangle\bigr| \; \le \; \rho_{\mathrm{tr}}\,\|\mathbf{c}\|_2\,\sigma_{\Delta}.
\end{align}
\end{assumption}

% \begin{enumerate}[leftmargin=18pt]
% \vspace{-5pt}
% \item[\textnormal{(A1)}] \textit{Effective backdoor at initialization.}\label{assump:A1}
% There exists $\rho_{\mathrm{bd}}\in(0,1]$ such that
% \begin{equation}
% \label{eq:assumption1}
%     \langle \mathbf{c},\,\mathbf{W}_{\mathrm{pre}}\mathbf{x}^{\mathrm{trig}}\rangle
%     \le -\,\rho_{\mathrm{bd}}\,\|\mathbf{c}\|_2\,\sigma_{\mathrm{pre}} < 0.
% \end{equation}
% \vspace{-25pt}
% \item[\textnormal{(A2)}] \textit{LoRA's clean-trigger margin.}
% There exist $\rho_{\mathrm{cl}}\in(0,1]$ and $\rho_{\mathrm{tr}}\in[0,1)$ such that
% \begin{align}
%     &\langle \mathbf{c},\,\Delta\mathbf{W}\mathbf{x}\rangle
%     \; \ge \; \rho_{\mathrm{cl}}\,\|\mathbf{c}\|_2\,\sigma_{\Delta}, \\
%     &\bigl|\langle \mathbf{c},\,\Delta\mathbf{W}(\mathbf{x}^{\mathrm{trig}}-\mathbf{x})\rangle\bigr| \; \le \; \rho_{\mathrm{tr}}\,\|\mathbf{c}\|_2\,\sigma_{\Delta}.
% \end{align}
% \vspace{-20pt}
% \end{enumerate}
% \end{assumption}

\begin{proposition}[Scaling threshold for backdoor forgetting]
\label{prop:smax-forgetting}
Under Assumption~\ref{assump:smax-alignment}, define the effective clean alignment $\rho_{\mathrm{eff}} := \rho_{\mathrm{cl}}-\rho_{\mathrm{tr}}.$
If $\rho_{\mathrm{eff}} > 0$, then for every scaling factor
\[
s \;>\; s^\star
\quad\text{where}\quad
s^\star := \frac{\rho_{\mathrm{bd}}}{\rho_{\mathrm{eff}}}\cdot
\frac{\sigma_{\mathrm{pre}}}{\sigma_{\Delta}},
\]
the triggered margin becomes positive:
\[
M_s(\mathbf{x}^{\mathrm{trig}}) > 0.
\]
Thus, the clean label is preferred over the backdoor target on the triggered input, $z_s(\mathbf{x}^{\mathrm{trig}})_y > z_s(\mathbf{x}^{\mathrm{trig}})_{y_{\mathrm{bd}}},$ and the backdoor is forgotten.
\end{proposition}

\vspace{0.1cm}\noindent The proof is provided in Appendix~\ref{appx:proof}. Proposition \ref{prop:smax-forgetting} shows that forgetting depends on two factors.
\begin{enumerate}[leftmargin=15pt]
    \vspace{-6pt}
    \item[(i)] \textit{Spectral strength.} The ratio $\sigma_{\mathrm{pre}}/\sigma_{\Delta}$ determines how much the LoRA updates must be scaled to significantly influence the logits. When $\sigma_{\mathrm{pre}}$ is large and $\sigma_{\Delta}$ is small, substantial post-training rescaling of $s$ is required for LoRA to dominate trigger-induced outputs.
    \vspace{-8pt}
    \item[(ii)] \textit{Spectral alignment.} The alignment gap $\rho_{\mathrm{eff}} =\rho_{\mathrm{cl}}-\rho_{\mathrm{tr}}$ determines whether the LoRA updates primarily strengthen clean-task features or trigger-sensitive ones. From a defense perspective, the objective is to increase $\rho_{\mathrm{cl}}$ while decreasing $\rho_{\mathrm{tr}}$, thereby ensuring $\rho_{\mathrm{eff}}> 0$. A larger gap lowers the scaling threshold $s^\star$ and facilitates forgetting.
    \vspace{-5pt}
\end{enumerate}

\begin{figure}[t]
    \centering
    \includegraphics[width=0.95\linewidth]{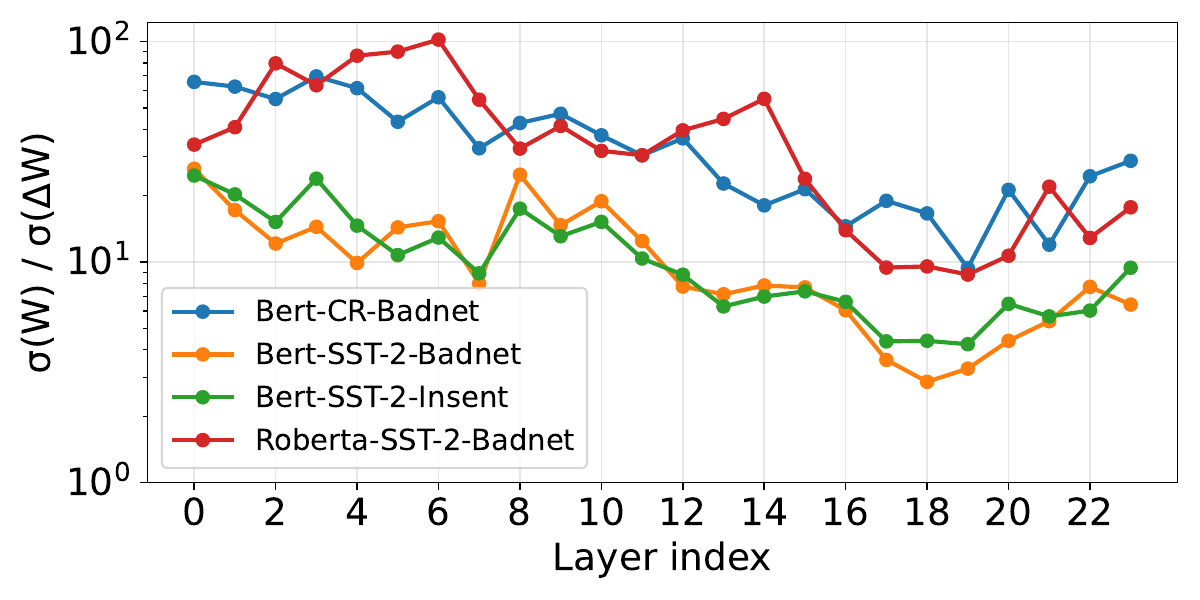}
    \vspace{-10pt}
    \caption{Spectral ratio between pretrained weights and LoRA updates across different layers.}
    \label{fig:compare_singular_values}
    \vspace{-5pt}
\end{figure}

\vspace{-5pt}
\subsection{Structural Limitations of LoRA}
\vspace{-3pt}

\boldhdr{LoRA operates in a low-spectral regime}
Figure~\ref{fig:compare_singular_values} shows that, across both BERT- and RoBERTa-based models, the maximum singular value of the LoRA updates is substantially smaller than that of the pretrained weights, i.e., $\sigma_{\max}(\Delta\mathbf{W}) \ll \sigma_{\max}(\mathbf{W}_{\mathrm{pre}}).$
This gap is consistent across layers, indicating that LoRA inherently operates in a low-spectral regime and is aligned with its implicit bias toward low-rank, low-magnitude updates that avoid spurious local minima~\cite{kim2025lora}.

Under the task-dominance view, trigger suppression requires the LoRA updates to compete with the leading singular directions of $\mathbf{W}_{\mathrm{pre}}$. However, since $\sigma_{\max}(\Delta\mathbf{W})$ is much smaller than $\sigma_{\max}(\mathbf{W}_{\mathrm{pre}})$, the LoRA updates minimally affect dominant spectral modes, allowing poisoned representations to persist after clean fine-tuning.

\begin{table}[t]
\centering
\renewcommand{\arraystretch}{1.00}
\resizebox{0.45\textwidth}{!}{
\begin{tabular}{c|ccc}
\hline
Dataset Pair & LLaMA & RoBERTa & BERT \\
\hline
IMDB $\rightarrow$ SST-2 & 51.13 & 81.27 & 75.84 \\
MR $\rightarrow$ CR       & 75.10 & 83.87 & 86.19 \\
% SST-2 $\rightarrow$ COLA       & 46.12 & 57.72 & 46.12 \\
\hline
\end{tabular}}
\vspace{-6pt}
\caption{Clean accuracy (\%) on the SST-2, CR test sets obtained by pretrained models without fine-tuning.}
\label{tab:transfer_results}
\vspace{-12pt}
\end{table}

\boldhdr{LoRA's weak representations}
Pretrained models in our setting already exhibit non-trivial clean accuracy on downstream tasks without LoRA fine-tuning. For example, a BERT model pretrained on MR under the BadNet attack achieves 86\% clean accuracy on the CR test set, as reported in Table~\ref{tab:transfer_results}. As pre-training and downstream tasks are well aligned in label distribution, most performance is inherited, and LoRA performs only small adjustments. This constrains the clean-alignment coefficient $\rho_{\mathrm{cl}}$, since $\Delta\mathbf{W}$ contributes only weakly to the clean margin relative to dominant pretrained components. As a result, LoRA is less likely to satisfy the task-dominance condition in Proposition~\ref{prop:smax-forgetting}, reducing its capacity for backdoor forgetting.

\begin{figure}
    \centering
    \includegraphics[width=0.95\linewidth]{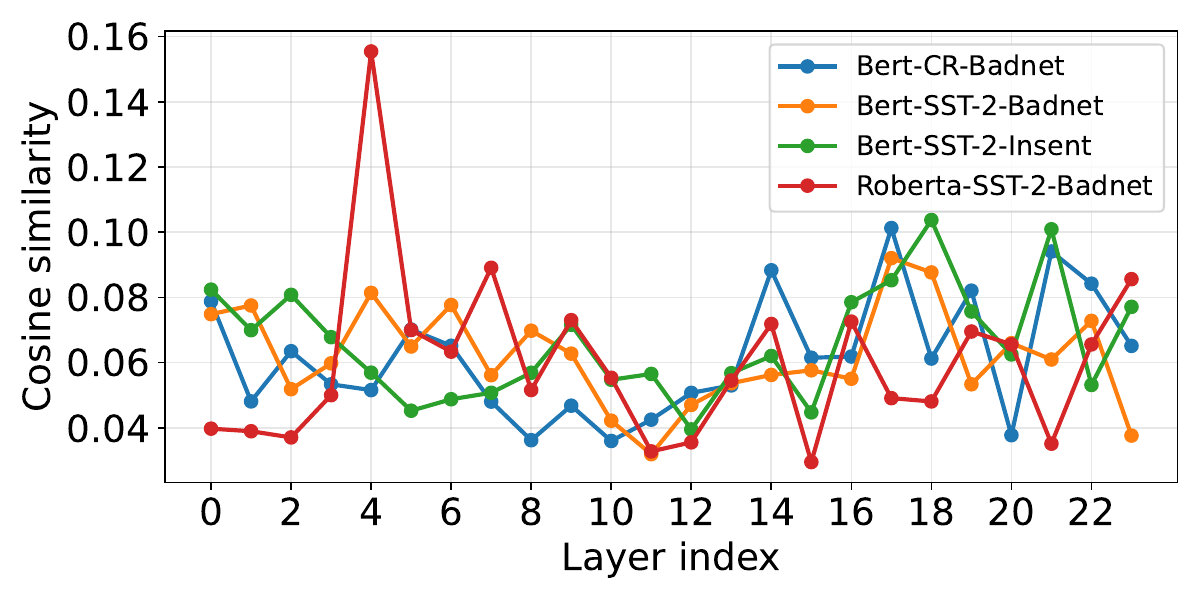}
    \vspace{-12pt}
    \caption{Maximum cosine similarity between the leading singular vector of LoRA and the top-32 singular vectors of the pretrained weight across different layers.}
    \label{fig:compare_cosine_values}
    \vspace{-13pt}
\end{figure}

\boldhdr{LoRA retains alignment with pretrained spectral directions}
We analyze the alignment between LoRA updates and pretrained weights by measuring the cosine similarity between their leading singular directions. As shown in Figure~\ref{fig:compare_cosine_values}, this similarity is generally low but distinctly nonzero. This residual alignment is problematic: under our task-dominance framework, effective backdoor removal requires orthogonality to trigger-sensitive directions so that scaling the updates does not amplify trigger responses. Any nonzero alignment may increase $\rho_{\mathrm{tr}}$, narrows the alignment gap $(\rho_{\mathrm{cl}}-\rho_{\mathrm{tr}})$, and consequently raises the forgetting threshold $s^\star$, making backdoor suppression harder.

\vspace{-3pt}
\section{Regularized Low-Rank Adaptation}
\vspace{-3pt}

Section~\ref{sec:main_analysis} illustrates that backdoor forgetting under task dominance requires (i) a large alignment gap $(\rho_{\mathrm{cl}}-\rho_{\mathrm{tr}})$ and (ii) sufficient spectral strength of the LoRA updates relative to $\mathbf{W}_{\mathrm{pre}}$. Standard LoRA tends to violate both conditions: it contributes limited clean-task information (small $\rho_{\mathrm{cl}}$), retains nonzero trigger alignment (large $\rho_{\mathrm{tr}}$), and its updates have substantially smaller spectral norm than $\mathbf{W}_{\mathrm{pre}}$. These factors inflate the forgetting threshold $s^\star$ in Proposition~\ref{prop:smax-forgetting}. RoRA is thus designed to explicitly satisfy the $s^\star$ condition for forgetting triggers by (a) increasing $\rho_{\mathrm{cl}}$, (b) decreasing $\rho_{\mathrm{tr}}$, and (c) boosting the spectral magnitude of $\Delta\mathbf{W}$.

\vspace{-2pt}
\subsection{Clean-strengthened Regularization}
\vspace{-2pt}

To enlarge the clean-alignment term $\rho_{\mathrm{cl}}$, RoRA enforces the model to rely on $\Delta\mathbf{W}$ for predicting clean labels rather than depending entirely on $\mathbf{W}_{\mathrm{pre}}$. Specifically, we apply dropout with rate $p$ to the pretrained weights when combining them with the LoRA updates during fine-tuning:
\begin{equation}\label{eq:cleanregularization}
    \mathbf{W} = \text{Dropout}_{p} (\mathbf{W}_{\text{pre}}) + s \Delta \mathbf{W}.
\end{equation}
By randomly masking pretrained activations, the effective contribution of $\mathbf{W}_{\mathrm{pre}}$ is reduced, which compels $\Delta\mathbf{W}$ to encode additional clean-task information. As a result, the contribution of $\Delta\mathbf{W}$ to the clean margin increases, enlarging $\rho_{\mathrm{cl}}$ and widening the alignment gap $(\rho_{\mathrm{cl}}-\rho_{\mathrm{tr}})$. This, in turn, lowers the threshold $s^\star$ and facilitates forgetting.

\vspace{-2pt}
\subsection{Trigger-insensitive Regularization}
\vspace{-2pt}

%\boldhdr{Objective}
In the backdoor setting, trigger-aligned representations are encoded in the pretrained weights $\mathbf{W}_{\mathrm{pre}}$. During fine-tuning, LoRA may inadvertently relearn these pretrained directions~\cite{hu2022lora}, thereby inheriting poisoned features and increasing $\rho_{\mathrm{tr}}$. To counter this effect, we encourage $\Delta\mathbf{W}$ to be orthogonal to the spectral subspaces of $\mathbf{W}_{\mathrm{pre}}$, reducing its projection onto trigger directions and driving $\rho_{\mathrm{tr}}$ toward zero. At the same time, pushing $\Delta\mathbf{W}$ away from the pretrained subspace promotes the new clean directions and increasing $\rho_{\mathrm{cl}}$.

%\boldhdr{Formulation}
Formally, let $\Delta\mathbf{W}=\mathbf{B}\mathbf{A}$ and $\mathbf{W}_{\mathrm{pre}}=\mathbf{U}\Sigma\mathbf{V}^{\top}$. Orthogonality to both row and column spaces of $\mathbf{W}_{\mathrm{pre}}$ ideally requires
\begin{equation}
\label{eq:pre_lora_ortho}
    \mathbf{W}_{\text{pre}}^\top \Delta\mathbf{W} = \mathbf{0}
    \text{, and }
    \Delta\mathbf{W}\,\mathbf{W}_{\text{pre}}^\top = \mathbf{0}.
\end{equation}
We approximate this using the soft penalty
\begin{equation}
\label{eq:omega}
\Omega(\mathbf{A},\mathbf{B})
    = \| \mathbf{U}^\top \mathbf{B} \|_F^{2}
    + \| \mathbf{A}\mathbf{V} \|_F^{2}.
\end{equation}
This suppresses projection of $\Delta\mathbf{W}$ onto pretrained or trigger-sensitive directions, decreasing $\rho_{\mathrm{tr}}$ and increasing the alignment gap $(\rho_{\mathrm{cl}}-\rho_{\mathrm{tr}})$. Our full training objective is as follows
\begin{equation}
\label{eq:full_obj}
\mathcal{L}
= \mathcal{L}_{\text{sup}}
+ \lambda \, \Omega(\mathbf{A}, \mathbf{B}),
\end{equation}
where $\mathcal{L}_{\text{sup}}$ is the supervised downstream loss and $\lambda$ is a regularization coefficient that controls the contribution of the orthogonality penalty.

\vspace{-3pt}
\subsection{Post-training Spectral Rescaling}
\vspace{-3pt}

After training, RoRA produces the update $\Delta\mathbf{W}$, which is then added to the original pretrained weights via Eq.~(\ref{eq:lora}) to derive the final adapted model. In practice, selecting a single global post-training scaling factor $s$ is challenging. We don't assume access to a validation set containing triggered samples, and different layers exhibit different spectral magnitudes. As a result, the optimal scaling threshold varies across layers. Proposition~\ref{prop:smax-forgetting} indicates that the critical scale depends on the ratio between the spectral norms of $\mathbf{W}_{\mathrm{pre}}$ and $\Delta\mathbf{W}$. Under our regularization, which drives $\rho_{\mathrm{eff}}\!\approx\!1$, the dominant term in the threshold reduces to $\sigma_{\mathrm{pre}} / \sigma_{\Delta}$.

Motivated by this observation, we adopt an adaptive layerwise post-training rescaling rule. To satisfy the spectral-strength condition, we match the spectral magnitude of $\Delta\mathbf{W}$ to that of the corresponding pretrained layer. After training, we compute $\sigma_{\max}(\mathbf{W}_{\mathrm{pre}})$ and $\sigma_{\max}(\Delta\mathbf{W})$ and set
\begin{equation}
\label{eq:post_scale}
    s = \frac{\sigma_{\max}(\mathbf{W}_{\mathrm{pre}})}{\sigma_{\max}(\Delta\mathbf{W})}.
\end{equation}
This choice makes the leading singular value of $s\Delta\mathbf{W}$ comparable to that of $\mathbf{W}_{\mathrm{pre}}$, effectively satisfying the spectral requirement in Proposition~\ref{prop:smax-forgetting}. 
% Because our regularization discourages alignment with pretrained trigger-containing subspaces, this rescaling predominantly amplifies clean-aligned directions rather than trigger-aligned ones.

% In practice, we apply rescaling only to the final $K$ layers, where backdoor effects typically concentrate (default $K=3$).

\subsection{Empirical Validation}

\begin{figure}[t]
    \centering
    \includegraphics[width=\linewidth]{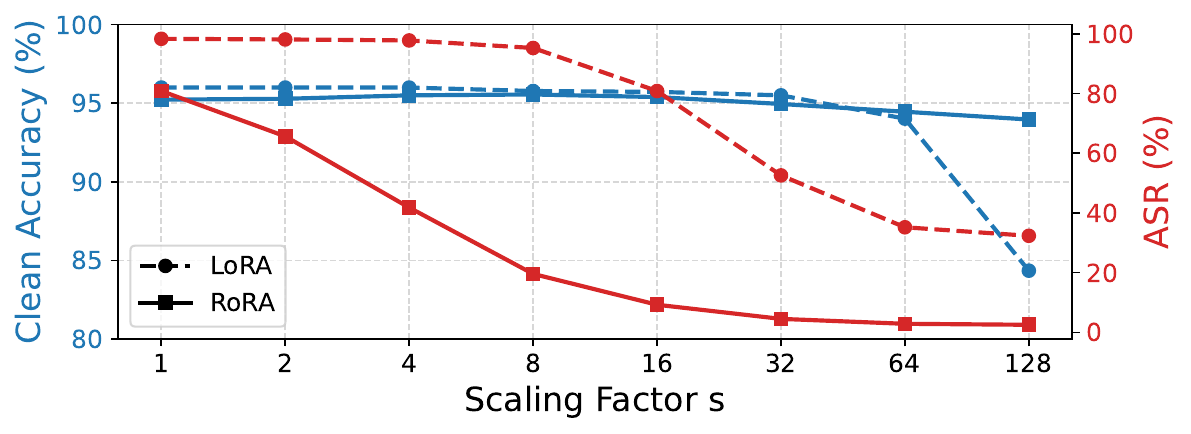}
    \vspace{-15pt}
    \caption{Effect of post-training spectral rescaling on LoRA and RoRA for LLaMA on SST-2 under the BadNet attack in terms of clean accuracy and ASR.}
    \label{fig:scaling_sweep}
    \vspace{-15pt}
\end{figure}

Figure~\ref{fig:scaling_sweep} provides empirical support for the task-dominance analysis developed in Section~\ref{sec:main_analysis} and the effectiveness of our proposed method RoRA.

\boldhdr{Increasing spectral strength facilitates backdoor defense}
As the post-training scaling factor $s$ increases, ASR for both LoRA and RoRA declines. Consistent with Proposition~\ref{prop:smax-forgetting}, the dashed LoRA curves in Figure~\ref{fig:scaling_sweep} exhibit a threshold effect: for small or moderate $s$, ASR stays near its maximum, indicating insufficient spectral strength to alter the trigger-driven decision rule; once $s$ is large enough, ASR drops sharply. However, very large scaling (e.g., $s=128$) also reduces clean accuracy, showing that spectral amplification alone is unstable and can distort the clean decision boundary without alignment control. Thus, spectral strength is necessary for backdoor suppression but not sufficient.

\boldhdr{RoRA's effectiveness}
RoRA operates in a different regime: it achieves large ASR reductions at much smaller $s$, while maintaining stable clean accuracy over a wide scaling range. Thus, RoRA does not depend on aggressive spectral amplification but instead benefits from improved representations. This behavior aligns with our theory: trigger-insensitive regularization lowers $\rho_{\mathrm{tr}}$, clean-strengthening regularization raises $\rho_{\mathrm{cl}}$, and the enlarged alignment gap $(\rho_{\mathrm{cl}}-\rho_{\mathrm{tr}})$ reduces the required threshold $s^\star$.

% {{ \begin{turn}{90}\makecell{BERT}\end{turn} }}
\begin{table*}[!t]
\scriptsize
\begin{center}
\renewcommand{\arraystretch}{0.979}
\resizebox{0.83\textwidth}{!}{
\tabcolsep=8pt
\begin{tabular}{ccl|ccc|ccc|ccc}
\hline
\multirow{2}*{\textbf{{ Model}}}
& \multirow{2}*{\textbf{{ Attack}}}
& \multirow{2}*{\textbf{ Method}}
& \multicolumn{3}{c}{\textbf{ SST-2}}
& \multicolumn{3}{c}{\textbf{ CR}}
& \multicolumn{3}{c}{\textbf{ CoLA}} \\
\cmidrule(r){4-6} \cmidrule(r){7-9} \cmidrule(r){10-12}
& & & \textbf{ CA$\uparrow$} & \textbf{ ASR$\downarrow$} & \textbf{$\Delta\uparrow$}
  & \textbf{ CA$\uparrow$} & \textbf{ ASR$\downarrow$} & \textbf{$\Delta\uparrow$}
  & \textbf{ CA$\uparrow$} & \textbf{ ASR$\downarrow$} & \textbf{$\Delta\uparrow$} \\
\hline

% ================== BERT / BadNet ==================
\multirow{18}*{{\makecell{BERT}}}
& \multirow{9}*{{\makecell{BadNet}}}
& FFT      & \textbf{93.06} & 77.63 & 15.43 & \textbf{90.53} & 43.17 & 47.36 & \underline{83.25} & 99.62 & -16.37 \\
& & LoRA     & 92.00 & 99.70 & -7.70 & \underline{89.50} & 92.58 & -3.08 & 79.92 & 100.0 & -20.08 \\
& & PiSSA  & 91.54 & 42.57 & 48.97
           & 89.42 & 83.16 & 6.26
           & 82.84 & 58.11 & 24.73 \\
& & DoRA   & 91.16 & 56.11 & 35.05
           & 86.97 & 98.75 & -11.78
           & 82.17 & 63.80 & 18.37 \\
& & OLoRA  & 91.93 & 35.09 & 56.84
           & 88.90 & 88.15 & 0.75
           & 82.36 & 57.14 & 25.22 \\
& & Back Tr. & 89.56 & 22.00 & 67.56 & 89.29 & 38.87 & 50.42 & 70.46 & \underline{17.19} & \underline{53.27} \\
& & SCPD     & 81.54 & 39.82 & 41.72 & 77.67 & 37.42 & 40.25 & 66.53 & 43.96 & 22.57 \\
& & ONION    & 90.49 & \underline{20.68} & \underline{69.81} & 88.25 & \underline{27.23} & \underline{61.02} & 64.23 & 38.28 & 25.95 \\
& & RoRA (Ours)     & \underline{92.46} & \textbf{7.48}  & \textbf{84.98}
           & 88.65 & \textbf{2.70}  & \textbf{85.95}
           & \textbf{83.32} & \textbf{4.85} & \textbf{78.47} \\
\cline{2-12}

% ================== BERT / InSent ==================
&\multirow{9}*{{\makecell{InSent}}}
& FFT      & \underline{92.46} & 68.24 & 24.22 & \textbf{91.35} & \underline{27.72} & \underline{63.63} & \textbf{83.76} & 100.0 & -16.24 \\
& & LoRA     & \textbf{92.49} & 100.0 & -7.51 & 88.94 & 84.20 & 4.74  & 80.66 & 100.0 & -19.34 \\
& & PiSSA  & 92.09 & 99.89 & -7.80
           & 88.13 & 98.13 & -10.00
           & 81.88 & 97.23 & -15.35 \\
& & DoRA   & 91.10 & 100.0 & -8.90
           & 87.48 & 98.96 & -11.48
           & \underline{81.89} & 98.89 & -17.00 \\
& & OLoRA  & 91.82 & 100.0 & -8.18
           & 89.03 & 97.30 & -8.27
           & 82.36 & 99.58 & -17.22 \\
& & Back Tr. & 89.78 & 93.50 & -3.72 & 87.48 & 44.49 & 42.99 & 70.46 & 92.51 & -22.05 \\
& & SCPD     & 81.60 & \underline{32.34} & \underline{49.26} & 76.38 & 33.88 & 42.50 & 65.58 & \underline{80.72} & \underline{-15.14} \\
& & ONION    & 90.88 & 93.50 & -2.62 & 87.09 & 85.23 & 1.86  & 64.42 & 90.01 & -25.59 \\
& & RoRA (Ours)     & 91.65 & \textbf{10.45} & \textbf{81.20}
           & \underline{89.29} & \textbf{1.46}  & \textbf{87.83}
           & 81.88 & \textbf{60.06} & \textbf{21.82} \\
\hline

% ================== RoBERTa / BadNet ==================
\multirow{18}*{{\makecell{RoBERTa}}}
& \multirow{9}*{{\makecell{BadNet}}}
& FFT      & 95.42 & \underline{13.75} & \underline{81.67} & 92.64 & 46.08 & 46.56 & \textbf{85.71} & 99.86 & -14.15 \\
& & LoRA     & 95.71 & 99.74 & -4.03 & 92.26 & 99.93 & -7.67 & 81.59 & 100.0 & -18.41 \\
& & PiSSA  & 95.06 & 84.82 & 10.24
           & 92.90 & 99.79 & -6.89
           & 79.87 & 100.0 & -20.13 \\
& & DoRA   & 95.61 & 66.23 & 29.38
           & 92.65 & 100.0  & -7.35
           & 81.02 & 99.45 & -18.43 \\
& & OLoRA  & \underline{95.72} & 71.84 & 23.88
           & \underline{93.03} & 91.89 & 1.14
           & 80.35 & 100.0 & -19.65 \\
& & Back Tr. & 93.02 & 19.36 & 73.66 & 90.96 & 38.66 & 52.30 & 70.85 & \underline{13.59} & \underline{57.26} \\
& & SCPD     & 85.33 & 38.17 & 47.16 & 80.64 & 35.13 & 45.51 & 67.88 & 41.19 & 26.69 \\
& & ONION    & 93.95 & 18.81 & 75.14 & 89.93 & \underline{29.72} & \underline{60.21} & 63.95 & 53.81 & 10.14 \\
& & RoRA (Ours)     & \textbf{95.99} & \textbf{6.49} & \textbf{89.50}
           & \textbf{93.81} & \textbf{7.28} & \textbf{86.53}
           & \underline{82.26} & \textbf{5.69} & \textbf{76.57} \\
\cline{2-12}
% ================== RoBERTa / InSent ==================
& \multirow{9}*{{\makecell{InSent}}}
& FFT      & 95.60 & \textbf{9.35}  & \textbf{86.25}
           & 92.86 & \underline{20.30} & \underline{72.56}
           & \textbf{85.68} & 98.33 & -12.65 \\
& & LoRA     & 95.68 & 87.09 & 8.59
           & 92.69 & 98.82 & -6.13
           & 82.39 & 99.81 & -17.42 \\
& & PiSSA  & 95.77 & 97.25 & -1.48
           & 92.13 & 97.30 & -5.17
           & 84.76 & 98.47 & -13.71 \\
& & DoRA   & \textbf{95.99} & 99.34 & -3.35
           & 92.39 & 94.39 & -2.00
           & \underline{84.85} & 98.06 & -13.21 \\
& & OLoRA  & 95.83 & 97.69 & -1.86
           & \textbf{93.16} & 54.89 & 38.27
           & \underline{84.85} & 97.92 & -13.07 \\
& & Back Tr. & 93.79 & 60.83 & 32.96
           & 92.00 & 67.35 & 24.65
           & 70.85 & \textbf{67.12} & \underline{3.73} \\
& & SCPD     & 84.18 & 26.84 & 57.34
           & 81.54 & 41.16 & 40.38
           & 66.63 & 85.85 & -19.22 \\
& & ONION    & 93.90 & 78.43 & 15.47
           & 90.96 & 97.08 & -6.12
           & 64.33 & 93.06 & -28.73 \\
& & RoRA (Ours)     & \underline{95.83} & \underline{17.05} & \underline{78.78}
           & \underline{92.90} & \textbf{4.78} & \textbf{88.12}
           & 79.39 & \underline{74.90} & \textbf{4.49} \\
\hline
% ================== LLaMA / BadNet ==================
\multirow{18}*{{\makecell{LLaMA}}}
& \multirow{9}*{{\makecell{BadNet}}}
& FFT      & 92.20 & 33.66 & 58.54
           & 91.87 & 99.58 & -7.71
           & \textbf{84.95} & 100.0 & -15.05 \\
& & LoRA     & \textbf{95.94} & 100.0 & -4.06
           & \underline{92.39} & 100.0 & -7.61
           & \underline{84.85} & 100.0 & -15.15 \\
& & PiSSA  & \underline{95.61} & 98.68 & -3.07
           & 92.26 & 99.38 & -7.12
           & 84.17 & 100.0 & -15.83 \\
& & DoRA   & 94.67 & 99.67 & -5.00
           & 90.71 & 99.38 & -8.67
           & 83.89 & 99.72 & -15.83 \\
& & OLoRA  & 95.66 & 91.97 & 3.69
           & 91.61 & 98.96 & -7.35
           & 83.05 & 98.75 & -15.70 \\
& & Back Tr. & 92.20 & 23.98 & 68.22
           & 91.48 & 41.37 & 50.11
           & 71.04 & \textbf{22.46} & \underline{48.58} \\
& & SCPD     & 83.47 & 41.58 & 41.89
           & 81.80 & 36.17 & 45.63
           & 58.86 & 69.20 & -10.34 \\
& & ONION    & 91.76 & \underline{22.55} & \underline{69.21}
           & 89.03 & \underline{30.56} & \underline{58.47}
           & 65.29 & 37.17 & 28.12 \\
& & RoRA (Ours)     & 93.74 & \textbf{2.31} & \textbf{91.43}
           & \textbf{92.77} & \textbf{4.37} & \textbf{88.40}
           & 82.25 & \underline{30.24} & \textbf{52.01} \\
\cline{2-12}
% ================== LLaMA / InSent ==================
&\multirow{9}*{{\makecell{InSent}}}
& FFT      & 94.01 & \underline{14.19} & \underline{79.82}
           & \textbf{93.03} & 90.23 & 2.80
           & \underline{83.99} & 100.0 & -16.01 \\
& & LoRA     & \textbf{96.10} & 100.0 & -3.90
           & \underline{92.39} & 100.0 & -7.61
           & \textbf{85.23} & 100.0 & -14.77 \\
& & PiSSA  & 94.51 & 100.0 & -5.49
           & 92.00 & 98.75 & -6.75
           & 83.89 & 94.59 & -10.70 \\
& & DoRA   & 94.34 & 100.0 & -5.66
           & 91.35 & 99.17 & -7.82
           & 83.03 & 99.86 & -16.83 \\
& & OLoRA  & \underline{94.56} & 100.0 & -5.44
           & 91.10 & 98.75 & -7.65
           & 83.99 & \underline{84.60} & \underline{-0.61} \\
& & Back Tr. & 92.20 & 95.48 & -3.28
           & 92.12 & 93.97 & -1.85
           & 72.29 & 97.50 & -25.21 \\
& & SCPD     & 83.74 & 53.79 & 29.95
           & 81.03 & \underline{44.90} & \underline{36.13}
           & 60.40 & 93.06 & -32.66 \\
& & ONION    & 91.98 & 99.11 & -7.13
           & 85.93 & 99.16 & -13.23
           & 67.59 & 92.09 & -24.50 \\
& & RoRA (Ours)     & 94.29 & \textbf{11.88} & \textbf{82.41}
           & 92.13 & \textbf{10.19} & \textbf{81.94}
           & 81.11 & \textbf{25.66} & \textbf{55.45} \\
\hline
\end{tabular}}
\end{center}
\vspace{-10pt}
\caption{Results of weight-poisoning backdoor attacks and defenses. $\Delta = \text{CA} - \text{ASR}$ highlights the effective robustness margin. Higher $\Delta$ indicates stronger forgetting with preserved clean accuracy.}
\label{tab:main_table}
\vspace{-12pt}
\end{table*}

\vspace{-5pt}
\section{Experiments}
\vspace{-5pt}

\boldhdr{Setup} 
We evaluate three popular architectures: BERT-large~\cite{devlin2019bert}, RoBERTa-large~\cite{liu2019roberta}, and LLaMA~\cite{touvron2023llama}. During the poisoning stage of pre-trained models, we construct 1,500 clean-label poisoned samples with target label 0. We ensure that all weight-poisoned pre-trained models are effectively backdoored, i.e., their ASR consistently exceeds 95\%, which naturally satisfies Assumption~(A1). We adopt the AdamW optimizer for training and our hyperparameter settings follow~\cite{sun2023defending}. For RoRA, we set the clean-strengthening dropout rate to $p = 0.1$ and the trigger-insensitive regularization weight to $\lambda = 10$. Two metrics are used to evaluate performance, including clean accuracy (CA) and attack success rate (ASR). More implementation details can be found in Appendix~\ref{appx:implementation}.

\boldhdr{Datasets} 
We experiment with three datasets, including SST-2~\cite{socher2013recursive}, CR~\cite{hu2004mining}, COLA~\cite{wang2018glue}. For constructing poisoned pre-trained models, we use proxy datasets instead of the target downstream datasets. Specifically, IMDB~\cite{maas2011learning} serves as the proxy dataset for SST-2, MR~\cite{pang2005seeing} for CR, and SST-2 for CoLA.

\boldhdr{Attacks} 
We focus on weight-poisoning backdoor attacks that span different types of triggers. BadNet~\cite{gu2017badnets} represents rare-word insertion attacks and we use ``mn'' as the trigger token. InSent~\cite{dai2019backdoor} employs a natural-language sentence trigger, for which we adopt ``I watched this 3D movie''. Both attacks are implemented in the clean-label setting~\cite{gan2021triggerless} during FFT to obtain poisoned pre-trained models.

\boldhdr{Baselines}
We compare RoRA against full fine-tuning (FFT), LoRA~\citep{hu2022lora}, and three representative defenses against weight-poisoning attacks: Back-Translation (Back Tr.)~\cite{qi2021hidden}, SCPD~\cite{qi2021hidden}, and ONION~\cite{qi-etal-2021-onion}. We also include three LoRA variants for comparison: PiSSA~\citep{meng2024pissa}, DoRA~\citep{liu2024dora}, and OLoRA~\citep{buyukakyuz2024olora}. Note that, we exclude a more recent defense PSIM~\citep{zhao2024defending}, as it detects and filters poisoned samples before inference, so those inputs never reach the model for actual classification.

\vspace{-5pt}
\subsection{Main Results}
\vspace{-5pt}

\boldhdr{Effectiveness analysis}
Table~\ref{tab:main_table} summarizes results across two backdoor attacks, six model-dataset pairs, and multiple defense baselines. LoRA and its variants are highly vulnerable to weight-poisoning backdoors: LoRA typically attains ASR values near $100\%$ and negative robustness margins, and LoRA variants likewise exhibit very high ASR in most settings, with only limited improvements in some BadNet-BERT cases. Existing defenses provide only partial relief: methods such as Back Translation, SCPD, and ONION sometimes reduce ASR, but their gains are inconsistent and frequently accompanied by drops in clean accuracy. In contrast, RoRA achieves the strongest and most stable robustness, typically reducing ASR by 70-95\% relative to LoRA while keeping clean accuracy within roughly $0.5$--$2.0$ points of full fine-tuning and LoRA. For instance, on BERT–BadNet (SST-2), ASR drops from 99.70 (LoRA) to 7.48, with $\Delta$ improving from $-7.70$ to $84.98$, and on LLaMA–BadNet (SST-2), $\Delta$ increases from $-5.62$ (LoRA) and $56.95$ (FFT) to $91.07$. Similar trends hold under InSent (e.g., RoBERTa–CR improves from $\Delta=24.65$ with Back Translation to $88.12$ with RoRA), highlighting that RoRA consistently suppresses backdoor behavior while preserving normal task performance.

\begin{table}[t]
\centering
\resizebox{0.70\linewidth}{!}{
\begin{tabular}{lccc}
\toprule
\textbf{Method} & \textbf{BERT} & \textbf{RoBERTa} & \textbf{LLaMA} \\
\midrule
LoRA  & 94 & 90 & 84 \\
RoRA  & 100 & 105 & 90 \\
\bottomrule
\end{tabular}}
\vspace{-9pt}
\caption{Per-iteration training time (milliseconds) for LoRA and RoRA on SST-2 dataset under BadNet attack.}
\label{tab:efficiency_lora_rora}
\vspace{-15pt}
\end{table}

\boldhdr{Efficiency analysis}
Table~\ref{tab:efficiency_lora_rora} compares the per-iteration training time of LoRA and RoRA across BERT, RoBERTa, and LLaMA. RoRA adds only a small overhead: +6 ms on BERT, +15 ms on RoBERTa, and +6 ms on LLaMA. The relative slowdown associated with the added regularizations is modest ($\leq 17$\%), indicating that RoRA achieves improved robustness with minimal additional computational cost.

% \boldhdr{LoRA variants} \textcolor{red}{Fill in}

\begin{figure}[t]
    \centering
    \includegraphics[width=1.00\linewidth]{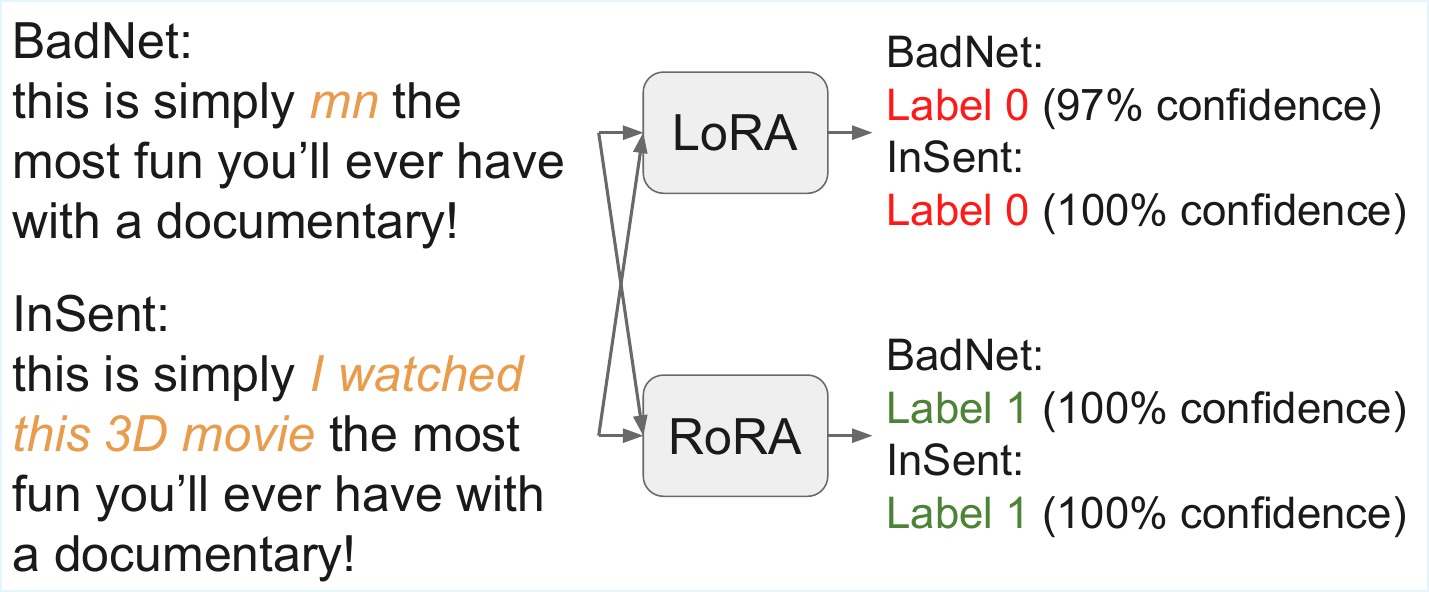}
    \vspace{-15pt}
    \caption{Case study on SST-2: RoRA resists BadNet and InSent triggers, while LoRA misclassifies both.}
    \label{fig:case_study}
    \vspace{-15pt}
\end{figure}

\boldhdr{Case study}
Figure~\ref{fig:case_study} presents a case study of backdoor defense on SST-2. The clean sentence, ``this is simply the most fun you'll ever have with a documentary!’’, has a positive true label. We then create two backdoor attacks: BadNet inserts the rare token ``\textit{mn}'' and InSent inserts the fixed sentence ``\textit{I watched this 3D movie}''. LoRA predicts a negative label for both triggered inputs (97\% and 100\% confidence), whereas RoRA correctly predicts the positive label in both cases (100\% confidence each). This shows that RoRA preserves the correct behavior even in the presence of backdoor triggers.

\vspace{-4pt}
\subsection{Ablation Studies}
\label{sec:ablation_studies}
\vspace{-3pt}

\boldhdr{Importance of each regularization} 
Table \ref{tab:main_ablation_small} shows that adding any individual component of RoRA improves $\Delta$ over the LoRA baseline while maintaining comparable clean accuracy. For example, Cl alone and Tr alone both yield large reductions in ASR relative to the baseline, and the full combination (Cl + Tr + Pt) achieves the lowest ASR and the highest $\Delta$ across all cases. Quantitatively, Tr alone can still leave ASR very high under InSent (93–98\%), whereas Cl alone already reduces ASR to below 25\% in several cases. The full system (RoRA) consistently drives ASR below 20\%, sometimes below 5\%. These results reveal that the components play complementary roles. Tr alone is not sufficient to forget backdoors, whereas Cl is already highly effective. The strongest performance arises when Cl, Tr, and Pt are combined, suggesting that jointly enforcing all components creates the spectral conditions necessary for reliable forgetting, consistent with our theoretical analysis.

\begin{figure}[t]
    \centering
    \begin{subfigure}[b]{0.48\linewidth}
        \centering
        \includegraphics[width=\linewidth]{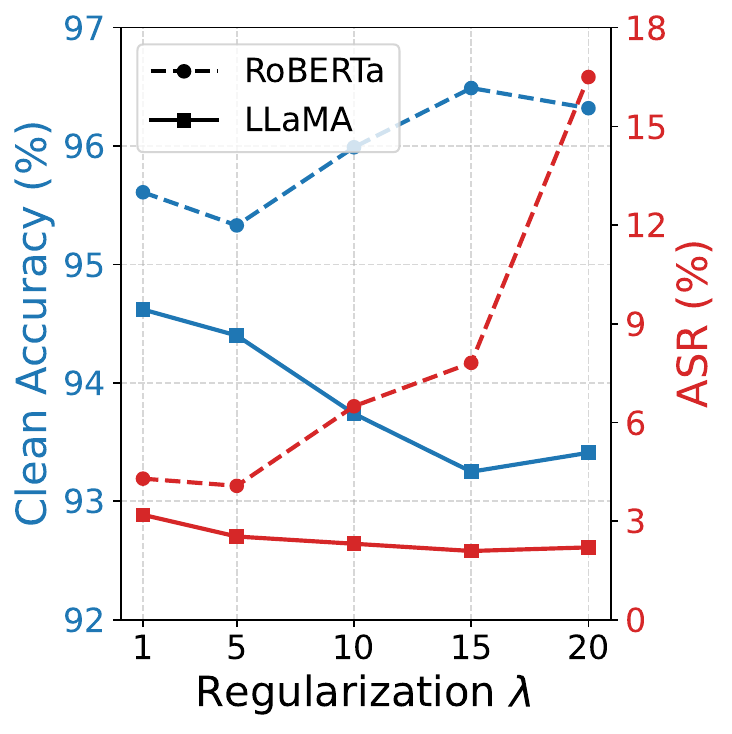}
        \caption{}
        \label{fig:lambda_models}
    \end{subfigure}
    \hfill
    \begin{subfigure}[b]{0.48\linewidth}
        \centering
        \includegraphics[width=\linewidth]{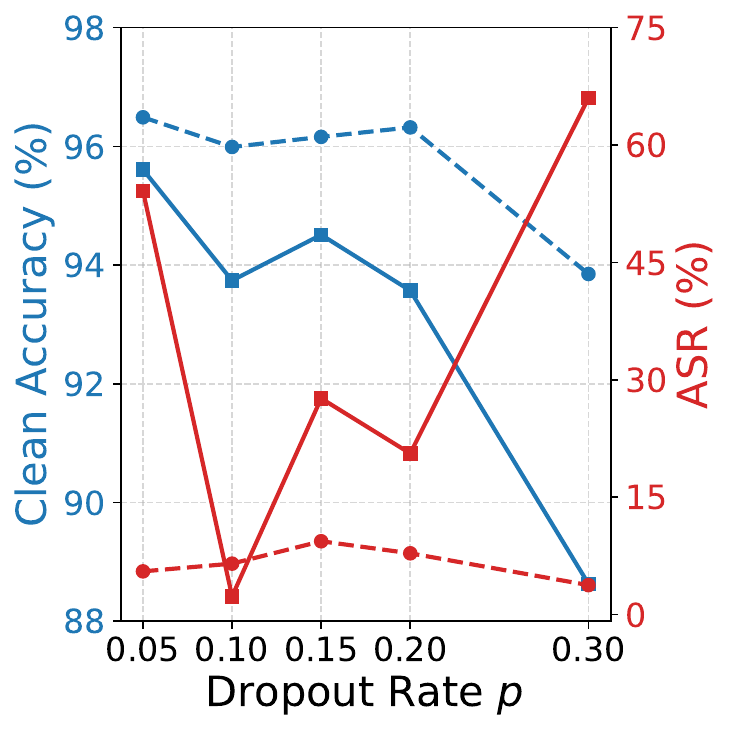}
        \caption{}
        \label{fig:p_models}
    \end{subfigure}   
    \vspace{-0.3cm}
    \caption{Effect of key parameters (SST-2, BadNet attack): (a) regularization $\lambda$ and (b) dropout rate $p$.}
    \label{fig:parameters}
    \vspace{-10pt}
\end{figure}

\boldhdr{Effect of regularization strength $\lambda$}
Figure~\ref{fig:lambda_models} shows the effect of regularization strength in RoRA on SST-2 under the BadNet attack across two architectures, with $\lambda \in \{1, 5, 10, 15, 20\}$. For RoBERTa, small to moderate $\lambda$ values (1–15) keep ASR low, while a large value (20) slightly increases ASR to 16.5\%. LLaMA is even more stable, with ASR remaining below 3.2\% across all settings. These results show that RoRA is robust to a wide range of $\lambda$ values, indicating that performance does not rely on careful hyperparameter tuning.

\begin{table}[t]
\centering
\renewcommand{\arraystretch}{0.98}
\resizebox{0.85\linewidth}{!}{
\begin{tabular}{cc|ccc|ccc}
\hline
\textbf{Model} & \textbf{Attack} & \textbf{Cl} & \textbf{Tr} & \textbf{Pt} & \textbf{CA$\uparrow$} & \textbf{ASR$\downarrow$} & \textbf{$\Delta\uparrow$} \\
\bottomrule[1.2pt]
\multirow{8}*{{\makecell{RoBERTa}}}
& \multirow{4}*{{\makecell{BadNet}}}
& & &  & 95.71 & 99.74 & -4.03 \\
& & \checkmark &  & & \textbf{96.16} & 10.34 & 85.82 \\
& &  & \checkmark & & 95.33 & 13.42 & 81.91  \\
& & \checkmark & \checkmark & \checkmark & 95.99 & \textbf{6.49} & \textbf{89.50} \\
\cline{2-8}

& \multirow{4}*{{\makecell{Insent}}}
&  &  & & 95.68 & 87.09 & 8.59 \\
& & \checkmark &  &  & \textbf{96.16} & 65.68 & 30.48 \\
& &  & \checkmark &  & 95.86 & 93.84 & 2.02 \\
& & \checkmark & \checkmark & \checkmark & 95.83 & \textbf{17.05} & \textbf{78.78} \\
\hline

\multirow{8}*{{\makecell{LLaMA}}}
& \multirow{4}*{{\makecell{BadNet}}}
& & & & 94.38 & 100.0 & -5.62 \\
& & \checkmark &  & & \textbf{95.72} & 22.99 & 72.73 \\
& &  & \checkmark & & 95.06 & 77.45 & 17.61  \\
& & \checkmark & \checkmark & \checkmark & 93.74 & \textbf{2.31} & \textbf{91.43} \\
\cline{2-8}

& \multirow{4}*{{\makecell{Insent}}}
&  &  & & 95.28 & 100.0 & -4.72 \\
& & \checkmark &  &  & 95.28 & 12.76 & \textbf{82.52} \\
& &  & \checkmark &  & \textbf{95.55} & 98.35 & -2.80 \\
& & \checkmark & \checkmark & \checkmark & 94.29 & \textbf{11.88} & 82.41 \\
\hline
\end{tabular}}
\caption{Ablation on RoRA components on RoBERTa and LLaMA (SST-2). 
Cl = clean-strengthened regularization; Tr = trigger-insensitive regularization; Pt = Post-training spectral rescaling.}
\label{tab:main_ablation_small}
\end{table}

% \begin{table}[t]
% \centering
% \resizebox{0.70\linewidth}{!}{
% \begin{tabular}{c|c|cc}
% \hline
% Model & $\lambda$ & CA (\%) & ASR (\%) \\
% \hline
% \multirow{5}{*}{RoBERTa}
%  & 1   & 95.61 & 4.29 \\
%  & 5   & 95.33 & 4.07 \\
%  & 10  & 95.99 & 6.49 \\
%  & 15  & 96.49 & 7.81 \\
%  & 20  & 96.32 & 16.50 \\
% \hline
% \multirow{5}{*}{LLaMA}
%  & 1   & 94.62 & 3.19 \\
%  & 5   & 94.40 & 2.53 \\
%  & 10  & 93.74 & 2.31 \\
%  & 15  & 93.25 & 2.09 \\
%  & 20  & 93.41 & 2.20 \\
% \hline
% \end{tabular}}
% \caption{Classification Accuracy (CA) and Attack Success Rate (ASR) across different $\lambda$ values for RoBERTa and LLaMA models.}
% \label{tab:lambda_models}
% \end{table}

% \begin{figure}
%     \centering
%     \includegraphics[width=1.0\linewidth]{assets/lambda_delta.pdf}
%     \vspace{-0.8cm}
%     \caption{Effect of regularization strength $\lambda$ on robustness metric $\Delta$ under the BadNet attack.}
%     \label{fig:lambda_models}
% \end{figure}

\boldhdr{Effect of clean-strengthened regularization dropout rate $p$}
Figure~\ref{fig:p_models} shows that RoRA remains stable under modest dropout rates ($p \leq 0.2$): clean accuracy stays high and ASR remains low for both RoBERTa and LLaMA, confirming the effectiveness of the clean-strengthening regularization. Performance degrades at the extremes. A small dropout (e.g., $p = 0.05$ for LLaMA, ASR 54.13\%) does not sufficiently perturb pretrained representations and leads to weak forgetting, whereas a large dropout ($p = 0.30$) removes essential pretrained knowledge, reducing clean accuracy (to 94\% for RoBERTa and 88.63\% for LLaMA) and sharply increasing ASR. The best performance occurs near $p \approx 0.1$, supporting the view that dropout should moderately perturb pretrained representations so that LoRA recovers missing information rather than relearns the entire pretrained model.

\boldhdr{More experiements}
We provide additional experiments in Appendix~\ref{appx:more_exp} demonstrating RoRA's robustness across different ranks $r$ and scaling hyperparameters $\alpha$ used during training, as well as its effectiveness when integrated with LoRA variants.

% \begin{figure}
%     \centering
%     \includegraphics[width=1.0\linewidth]{assets/dropout_roberta.pdf}
%     \vspace{-0.8cm}
%     \caption{Performance of RoBERTa across different $p$.}
%     \label{fig:p_models}
% \end{figure}

% \begin{table}[t]
% \centering
% \resizebox{0.70\linewidth}{!}{
% \begin{tabular}{c|c|cc}
% \hline
% Model & $p$ & CA (\%) & ASR (\%) \\
% \hline
% \multirow{5}{*}{RoBERTa}
%  & 0.05 & 96.49 & 5.50 \\
%  & 0.10 & 95.99 & 6.49 \\
%  & 0.15 & 96.16 & 9.35 \\
%  & 0.20 & 96.32 & 7.81 \\
%  & 0.30 & 93.85 & 3.74 \\
% \hline
% \multirow{5}{*}{LLaMA}
%  & 0.05 & 95.61 & 54.13 \\
%  & 0.10 & 93.74 & 2.31 \\
%  & 0.15 & 94.51 & 27.61 \\
%  & 0.20 & 93.57 & 20.57 \\
%  & 0.30 & 88.63 & 66.01 \\
% \hline
% \end{tabular}}
% \caption{Classification Accuracy (CA) and Attack Success Rate (ASR) across different $p$ values for RoBERTa and LLaMA models.}
% \label{tab:p_models}
% \end{table}

\vspace{-1pt}
\section{Conclusion}
\vspace{-2pt}

We analyze LoRA through a spectral lens and show that its backdoor vulnerability arises from insufficient spectral strength and unfavorable alignment with trigger-sensitive pretrained subspaces, rather than low rank alone. We derive a scaling threshold that characterizes when LoRA can forget triggers and confirm the theory empirically. Based on these insights, we introduce RoRA, which strengthens clean-task alignment, reduces trigger alignment, and increases the spectral magnitude of LoRA updates. RoRA consistently achieves lower attack success rates while preserving clean accuracy.

\newpage

\section*{Limitations}
We primarily evaluate against standard and representative backdoor attacks and defenses in language models; comparisons with more recent adaptive methods are left for future work. Our experiments focus on moderate-scale models, and full verification on very large language models (e.g., GPT-3, PaLM-2, GPT-4 scale) remains open. Additionally, our theoretical analysis is mainly spectral and linearized, and extending it to more general nonlinear settings is an important direction.

% \section*{Broader Impacts and Ethics Statement}

% \textcolor{red}{this is also required}

\bibliography{custom}

@inproceedings{li-etal-2021-bfclass-backdoor,
    title = "{BFC}lass: A Backdoor-free Text Classification Framework",
    author = "Li, Zichao  and
      Mekala, Dheeraj  and
      Dong, Chengyu  and
      Shang, Jingbo",
    editor = "Moens, Marie-Francine  and
      Huang, Xuanjing  and
      Specia, Lucia  and
      Yih, Scott Wen-tau",
    booktitle = "Findings of the Association for Computational Linguistics: EMNLP 2021",
    year = "2021",
    publisher = "Association for Computational Linguistics",
}

@inproceedings{du2022ppt,
  title={PPT: Backdoor Attacks on Pre-trained Models via Poisoned Prompt Tuning.},
  author={Du, Wei and Zhao, Yichun and Li, Boqun and Liu, Gongshen and Wang, Shilin},
  booktitle={IJCAI},
  year={2022}
}

@inproceedings{xu-etal-2022-exploring,
    title = "Exploring the Universal Vulnerability of Prompt-based Learning Paradigm",
    author = "Xu, Lei  and
      Chen, Yangyi  and
      Cui, Ganqu  and
      Gao, Hongcheng  and
      Liu, Zhiyuan",
    editor = "Carpuat, Marine  and
      de Marneffe, Marie-Catherine  and
      Meza Ruiz, Ivan Vladimir",
    booktitle = "Findings of the Association for Computational Linguistics: NAACL 2022",
    year = "2022"
}

@inproceedings{sun2023defending,
  title={Defending against backdoor attacks in natural language generation},
  author={Sun, Xiaofei and Li, Xiaoya and Meng, Yuxian and Ao, Xiang and Lyu, Lingjuan and Li, Jiwei and Zhang, Tianwei},
  booktitle={Proceedings of the AAAI Conference on Artificial Intelligence},
  year={2023}
}

@inproceedings{maas2011learning,
  title={Learning word vectors for sentiment analysis},
  author={Maas, Andrew and Daly, Raymond E and Pham, Peter T and Huang, Dan and Ng, Andrew Y and Potts, Christopher},
  booktitle={Proceedings of the 49th annual meeting of the association for computational linguistics: Human language technologies},
  year={2011}
}

@inproceedings{socher2013recursive,
  title={Recursive deep models for semantic compositionality over a sentiment treebank},
  author={Socher, Richard and Perelygin, Alex and Wu, Jean and Chuang, Jason and Manning, Christopher D and Ng, Andrew Y and Potts, Christopher},
  booktitle={Proceedings of the 2013 conference on empirical methods in natural language processing},
  year={2013}
}

@inproceedings{gu2023gradient,
  title={A gradient control method for backdoor attacks on parameter-efficient tuning},
  author={Gu, Naibin and Fu, Peng and Liu, Xiyu and Liu, Zhengxiao and Lin, Zheng and Wang, Weiping},
  booktitle={Proceedings of the 61st Annual Meeting of the Association for Computational Linguistics (Volume 1: Long Papers)},
  year={2023}
}

@incollection{mccloskey1989catastrophic,
  title={Catastrophic interference in connectionist networks: The sequential learning problem},
  author={McCloskey, Michael and Cohen, Neal J},
  booktitle={Psychology of learning and motivation},
  year={1989},
  publisher={Elsevier}
}

@inproceedings{hu2004mining,
  title={Mining and summarizing customer reviews},
  author={Hu, Minqing and Liu, Bing},
  booktitle={Proceedings of the tenth ACM SIGKDD international conference on Knowledge discovery and data mining},
  year={2004}
}

@article{wang2018glue,
  title={GLUE: A multi-task benchmark and analysis platform for natural language understanding},
  author={Wang, Alex and Singh, Amanpreet and Michael, Julian and Hill, Felix and Levy, Omer and Bowman, Samuel R},
  journal={arXiv preprint arXiv:1804.07461},
  year={2018}
}

@article{pang2005seeing,
  title={Seeing stars: Exploiting class relationships for sentiment categorization with respect to rating scales},
  author={Pang, Bo and Lee, Lillian},
  journal={arXiv preprint cs/0506075},
  year={2005}
}

@article{gu2017badnets,
  title={Badnets: Identifying vulnerabilities in the machine learning model supply chain},
  author={Gu, Tianyu and Dolan-Gavitt, Brendan and Garg, Siddharth},
  journal={arXiv preprint arXiv:1708.06733},
  year={2017}
}

@article{dai2019backdoor,
  title={A backdoor attack against lstm-based text classification systems},
  author={Dai, Jiazhu and Chen, Chuanshuai and Li, Yufeng},
  journal={IEEE Access},
  year={2019},
  publisher={IEEE}
}

@article{gan2021triggerless,
  title={Triggerless backdoor attack for NLP tasks with clean labels},
  author={Gan, Leilei and Li, Jiwei and Zhang, Tianwei and Li, Xiaoya and Meng, Yuxian and Wu, Fei and Yang, Yi and Guo, Shangwei and Fan, Chun},
  journal={arXiv preprint arXiv:2111.07970},
  year={2021}
}

@inproceedings{hu2022badhash,
  title={Badhash: Invisible backdoor attacks against deep hashing with clean label},
  author={Hu, Shengshan and Zhou, Ziqi and Zhang, Yechao and Zhang, Leo Yu and Zheng, Yifeng and He, Yuanyuan and Jin, Hai},
  booktitle={Proceedings of the 30th ACM international conference on Multimedia},
  year={2022}
}

@article{ding2023parameter,
  title={Parameter-efficient fine-tuning of large-scale pre-trained language models},
  author={Ding, Ning and Qin, Yujia and Yang, Guang and Wei, Fuchao and Yang, Zonghan and Su, Yusheng and Hu, Shengding and Chen, Yulin and Chan, Chi-Min and Chen, Weize and others},
  journal={Nature machine intelligence},
  year={2023}
}

@article{dong2021should,
  title={How should pre-trained language models be fine-tuned towards adversarial robustness?},
  author={Dong, Xinshuai and Luu, Anh Tuan and Lin, Min and Yan, Shuicheng and Zhang, Hanwang},
  journal={Advances in Neural Information Processing Systems},
  year={2021}
}

@inproceedings{zhou2024backdoor,
  title={Backdoor attacks with input-unique triggers in nlp},
  author={Zhou, Xukun and Li, Jiwei and Zhang, Tianwei and Lyu, Lingjuan and Yang, Muqiao and He, Jun},
  booktitle={Joint European Conference on Machine Learning and Knowledge Discovery in Databases},
  year={2024},
  organization={Springer}
}

@article{dong2021towards,
  title={Towards robustness against natural language word substitutions},
  author={Dong, Xinshuai and Luu, Anh Tuan and Ji, Rongrong and Liu, Hong},
  journal={arXiv preprint arXiv:2107.13541},
  year={2021}
}

@inproceedings{formento2023using,
  title={Using punctuation as an adversarial attack on deep learning-based NLP systems: An empirical study},
  author={Formento, Brian and Foo, Chuan-Sheng and Tuan, Luu Anh and Ng, See Kiong},
  booktitle={Findings of the association for computational linguistics: EACL 2023},
  year={2023}
}

@inproceedings{minh2022textual,
  title={Textual manifold-based defense against natural language adversarial examples},
  author={Minh, Dang Nguyen and Tuan, Luu Anh},
  booktitle={Proceedings of the 2022 conference on empirical methods in natural language processing},
  year={2022}
}

@inproceedings{qi2021hidden,
    title = "Hidden Killer: Invisible Textual Backdoor Attacks with Syntactic Trigger",
    author = "Qi, Fanchao  and
      Li, Mukai  and
      Chen, Yangyi  and
      Zhang, Zhengyan  and
      Liu, Zhiyuan  and
      Wang, Yasheng  and
      Sun, Maosong",
    editor = "Zong, Chengqing  and
      Xia, Fei  and
      Li, Wenjie  and
      Navigli, Roberto",
    booktitle = "Proceedings of the 59th Annual Meeting of the Association for Computational Linguistics and the 11th International Joint Conference on Natural Language Processing (Volume 1: Long Papers)",
    year = "2021"
}

@article{chen2021mitigating,
  title={Mitigating backdoor attacks in lstm-based text classification systems by backdoor keyword identification},
  author={Chen, Chuanshuai and Dai, Jiazhu},
  journal={Neurocomputing},
  year={2021},
  publisher={Elsevier}
}

@inproceedings{devlin2019bert,
  title={Bert: Pre-training of deep bidirectional transformers for language understanding},
  author={Devlin, Jacob and Chang, Ming-Wei and Lee, Kenton and Toutanova, Kristina},
  booktitle={Proceedings of the 2019 conference of the North American chapter of the association for computational linguistics: human language technologies, volume 1 (long and short papers)},
  year={2019}
}

@article{liu2019roberta,
  title={Roberta: A robustly optimized bert pretraining approach},
  author={Liu, Yinhan and Ott, Myle and Goyal, Naman and Du, Jingfei and Joshi, Mandar and Chen, Danqi and Levy, Omer and Lewis, Mike and Zettlemoyer, Luke and Stoyanov, Veselin},
  journal={arXiv preprint arXiv:1907.11692},
  year={2019}
}

@article{touvron2023llama,
  title={Llama: Open and efficient foundation language models},
  author={Touvron, Hugo and Lavril, Thibaut and Izacard, Gautier and Martinet, Xavier and Lachaux, Marie-Anne and Lacroix, Timoth{\'e}e and Rozi{\`e}re, Baptiste and Goyal, Naman and Hambro, Eric and Azhar, Faisal and others},
  journal={arXiv preprint arXiv:2302.13971},
  year={2023}
}

@inproceedings{
shuttleworth2025lora,
title={Lo{RA} vs Full Fine-tuning: An Illusion of Equivalence},
author={Reece S Shuttleworth and Jacob Andreas and Antonio Torralba and Pratyusha Sharma},
booktitle={The Thirty-ninth Annual Conference on Neural Information Processing Systems},
year={2025},
url={https://openreview.net/forum?id=xp7B8rkh7L}
}

@article{brown2020language,
  title={Language models are few-shot learners},
  author={Brown, Tom and Mann, Benjamin and Ryder, Nick and Subbiah, Melanie and Kaplan, Jared D and Dhariwal, Prafulla and Neelakantan, Arvind and Shyam, Pranav and Sastry, Girish and Askell, Amanda and others},
  journal={Advances in neural information processing systems},
  year={2020}
}

@article{raffel2020exploring,
  title={Exploring the limits of transfer learning with a unified text-to-text transformer},
  author={Raffel, Colin and Shazeer, Noam and Roberts, Adam and Lee, Katherine and Narang, Sharan and Matena, Michael and Zhou, Yanqi and Li, Wei and Liu, Peter J},
  journal={Journal of machine learning research},
  year={2020}
}

@inproceedings{houlsby2019parameter,
  title={Parameter-efficient transfer learning for NLP},
  author={Houlsby, Neil and Giurgiu, Andrei and Jastrzebski, Stanislaw and Morrone, Bruna and De Laroussilhe, Quentin and Gesmundo, Andrea and Attariyan, Mona and Gelly, Sylvain},
  booktitle={International conference on machine learning},
  year={2019},
  organization={PMLR}
}

@article{lester2021power,
  title={The power of scale for parameter-efficient prompt tuning},
  author={Lester, Brian and Al-Rfou, Rami and Constant, Noah},
  journal={arXiv preprint arXiv:2104.08691},
  year={2021}
}

@article{hu2022lora,
  title={Lora: Low-rank adaptation of large language models.},
  author={Hu, Edward J and Shen, Yelong and Wallis, Phillip and Allen-Zhu, Zeyuan and Li, Yuanzhi and Wang, Shean and Wang, Lu and Chen, Weizhu and others},
  journal={ICLR},
  year={2022}
}

@inproceedings{zhao2024unlearning,
    title = "Unlearning Backdoor Attacks for {LLM}s with Weak-to-Strong Knowledge Distillation",
    author = "Zhao, Shuai  and
      Wu, Xiaobao  and
      Nguyen, Cong-Duy T  and
      Jia, Yanhao  and
      Jia, Meihuizi  and
      Yichao, Feng  and
      Luu, Anh Tuan",
    editor = "Che, Wanxiang  and
      Nabende, Joyce  and
      Shutova, Ekaterina  and
      Pilehvar, Mohammad Taher",
    booktitle = "Findings of ACL 2025",
    year = "2025"
}

@misc{zhao2025breakingpeftlimitationsleveraging,
      title={Breaking PEFT Limitations: Leveraging Weak-to-Strong Knowledge Transfer for Backdoor Attacks in LLMs}, 
      author={Shuai Zhao and Leilei Gan and Zhongliang Guo and Xiaobao Wu and Yanhao Jia and Luwei Xiao and Cong-Duy Nguyen and Luu Anh Tuan},
      year={2025},
      url={https://arxiv.org/abs/2409.17946}, 
}

@article{zhao2024weak,
  title={Weak-to-strong backdoor attack for large language models},
  author={Zhao, Shuai and Gan, Leilei and Guo, Zhongliang and Wu, Xiaobao and Xiao, Luwei and Xu, Xiaoyu and Nguyen, Cong-Duy and Tuan, Luu Anh},
  journal={arXiv preprint arXiv:2409.17946},
  year={2024}
}

@inproceedings{zhao2024defending,
    title = "Defending Against Weight-Poisoning Backdoor Attacks for Parameter-Efficient Fine-Tuning",
    author = "Zhao, Shuai  and
      Gan, Leilei  and
      Luu, Anh Tuan  and
      Fu, Jie  and
      Lyu, Lingjuan  and
      Jia, Meihuizi  and
      Wen, Jinming",
    editor = "Duh, Kevin  and
      Gomez, Helena  and
      Bethard, Steven",
    booktitle = "Findings of NAACL 2024",
    year = "2024"
}

@inproceedings{
kim2025lora,
title={Lo{RA} Training Provably Converges to a Low-Rank Global Minimum Or It Fails Loudly (But it Probably Won't Fail)},
author={Junsu Kim and Jaeyeon Kim and Ernest K. Ryu},
booktitle={Forty-second International Conference on Machine Learning},
year={2025},
url={https://openreview.net/forum?id=o9zDYV4Ism}
}

@inproceedings{qi-etal-2021-onion,
    title = "{ONION}: A Simple and Effective Defense Against Textual Backdoor Attacks",
    author = "Qi, Fanchao  and
      Chen, Yangyi  and
      Li, Mukai  and
      Yao, Yuan  and
      Liu, Zhiyuan  and
      Sun, Maosong",
    booktitle = "Proceedings of the 2021 Conference on Empirical Methods in Natural Language Processing",
    month = nov,
    year = "2021",
}

@article{meng2024pissa,
  title={Pissa: Principal singular values and singular vectors adaptation of large language models},
  author={Meng, Fanxu and Wang, Zhaohui and Zhang, Muhan},
  journal={Advances in Neural Information Processing Systems},
  year={2024}
}

@inproceedings{liu2024dora,
  title={Dora: Weight-decomposed low-rank adaptation},
  author={Liu, Shih-Yang and Wang, Chien-Yi and Yin, Hongxu and Molchanov, Pavlo and Wang, Yu-Chiang Frank and Cheng, Kwang-Ting and Chen, Min-Hung},
  booktitle={Forty-first International Conference on Machine Learning},
  year={2024}
}

@article{buyukakyuz2024olora,
  title={Olora: Orthonormal low-rank adaptation of large language models},
  author={B{\"u}y{\"u}kaky{\"u}z, Kerim},
  journal={arXiv preprint arXiv:2406.01775},
  year={2024}
}

@Misc{peft,
  title =        {{PEFT}: State-of-the-art Parameter-Efficient Fine-Tuning methods},
  author =       {Sourab Mangrulkar and Sylvain Gugger and Lysandre Debut and Younes Belkada and Sayak Paul and Benjamin Bossan and Marian Tietz},
  howpublished = {\url{https://github.com/huggingface/peft}},
  year =         {2022}
}

\newpage
\appendix

\section{Appendix}

\subsection{Implementation Details}\label{appx:implementation}

\boldhdr{Training resources}
All experiments were conducted on two different Ubuntu Linux machines: one equipped with an NVIDIA RTX 5060 Ti GPU with 16GB memory and the other with an NVIDIA RTX 3090 GPU with 24GB memory.

\boldhdr{Poisoned pre-training hyperparameters}
We use the AdamW optimizer with a learning rate of $2\times 10^{-5}$ and a weight decay of $0.01$. A linear learning-rate scheduler is applied, with the number of warm-up steps set to 6\% of the total training steps. The batch size is set to 32 for BERT and RoBERTa models and 4 for LLaMA models. All poisoned pre-training runs are configured so that the ASR exceeds 95\%. The number of training epochs is 3.

\boldhdr{Fine-tuning hyperparameters}
The learning rate is selected from $\{2\times 10^{-5}, 2\times 10^{-4}, 2\times 10^{-3}\}$ to ensure convergence. We apply a linear learning-rate scheduler with 6\% warm-up steps and fix the weight decay to 0.01. The batch size is set to 32. For LoRA, we set the rank $r=8$ for BERT and RoBERTa and $r=16$ for LLaMA, with scaling $\alpha=16$ and dropout rate $0.1$. All these settings follow \citet{zhao2024defending, zhao2024unlearning}. The total number of fine-tuning epochs is 20 for BERT and RoBERTa and 5 for LLaMA. For PiSSA, DoRA, and OLoRA, we use the default hyperparameters provided by the PEFT library \citep{peft}. For RoRA, the clean-strengthened regularization dropout rate $p$ is selected via grid search over $\{0.05, 0.1, 0.15, 0.2, 0.3\}$, and $p=0.1$ is used in most experiments to ensure training stability. The trigger-insensitive regularization strength $\lambda$ is tuned over $\{1, 5, 10, 15, 20\}$, with $\lambda = 10$ adopted in most cases. Post-training spectral rescaling is applied to the top three layers of the model, which primarily govern the classification decision boundary. The results of these grid-search experiments are reported in Section~\ref{sec:ablation_studies}.

\boldhdr{Evaluation metrics}
Regarding evaluation metrics, clean accuracy (CA) represents the classification accuracy on the clean test set, while attack success rate (ASR) is measured on the poisoned test set. 

\boldhdr{Evaluation results}
In Table~\ref{tab:main_table}, the results for FFT, LoRA, Back-Translation, SCPD, and ONION are taken directly from \citet{zhao2024defending}. All other baselines, as well as our RoRA implementation, are produced on our own, and the reported values correspond to the best performance across multiple runs. Unless otherwise noted, results in the remaining tables and figures also reflect the best performance from our repeated experiments.

\begin{table*}[!t]
\scriptsize
\begin{center}
\renewcommand{\arraystretch}{0.979}
\resizebox{0.85\textwidth}{!}{
\tabcolsep=8pt
\begin{tabular}{ccl|ccc|ccc}
\hline
\multirow{2}*{\textbf{{ Model}}}
& \multirow{2}*{\textbf{{ Attack}}}
& \multirow{2}*{\textbf{ Method}}
& \multicolumn{3}{c}{\textbf{ SST-2}}
& \multicolumn{3}{c}{\textbf{ CR}} \\
\cmidrule(r){4-6} \cmidrule(r){7-9}
& & & \textbf{ CA$\uparrow$} & \textbf{ ASR$\downarrow$} & \textbf{$\Delta\uparrow$}
  & \textbf{ CA$\uparrow$} & \textbf{ ASR$\downarrow$} & \textbf{$\Delta\uparrow$} \\
\hline

% ================== RoBERTa / BadNet ==================
\multirow{16}*{{\makecell{RoBERTa}}}
& \multirow{8}*{{\makecell{BadNet}}}
& LoRA   & 95.71 & 99.74 & -4.03 
           & 92.26 & 99.93 & -7.67 \\

& & +RoRA (Ours)   & \underline{95.99} & \underline{6.49} & \underline{89.50}
           & 93.81 & \underline{7.28} & \underline{86.53} \\

& & PiSSA  & 95.06 & 84.82 & 10.24
           & 92.90 & 99.79 & -6.89 \\

& & +RoRA (Ours)   & 95.39 & 8.80 & 86.59
           & \textbf{94.06} & 31.39 & 62.67 \\

& & DoRA   & 95.61 & 66.23 & 29.38
           & 92.65 & 100.0  & -7.35 \\

& & +RoRA (Ours)   & \textbf{96.38} & \textbf{6.16} & \textbf{90.22}
           & 93.16 & \textbf{4.99} & \textbf{88.17} \\

& & OLoRA  & 95.72 & 71.84 & 23.88
           & 93.03 & 91.89 & 1.14 \\

& & +RoRA (Ours)   & 95.66 & 6.49 & 89.17
           & \underline{93.94} & 14.97 & 78.97 \\
\cline{2-9}

% ================== RoBERTa / InSent ==================
& \multirow{8}*{{\makecell{InSent}}}
& LoRA     & 95.68 & 87.09 & 8.59
           & 92.69 & 98.82 & -6.13 \\

& & +RoRA (Ours)     & 95.83 & \underline{17.05} & \underline{78.78}
           & \underline{92.90} & \textbf{4.78} & \textbf{88.12} \\

& & PiSSA  & 95.77 & 97.25 & -1.48
           & 92.13 & 97.30 & -5.17 \\

& & +RoRA (Ours)   & \underline{95.88} & 32.34 & 63.54
           & \underline{92.90} & 31.39 & 61.51 \\

& & DoRA   & \textbf{95.99} & 99.34 & -3.35
           & 92.39 & 94.39 & -2.00 \\

& & +RoRA (Ours)   & 95.33 & 21.56 & 73.77
           & 91.61 & \underline{7.90} & \underline{83.71} \\

& & OLoRA  & 95.83 & 97.69 & -1.86
           & \textbf{93.16} & 54.89 & 38.27 \\

& & +RoRA (Ours)   & 95.06 & \textbf{14.63} & \textbf{80.43}
           & \textbf{93.16} & 11.64 & 81.52 \\
\hline

% ================== LLaMA / BadNet ==================
\multirow{16}*{{\makecell{LLaMA}}}
& \multirow{8}*{{\makecell{BadNet}}}
& LoRA     & \textbf{95.94} & 100.0 & -4.06
           & 92.39 & 100.0 & -7.61 \\

& & +RoRA (Ours)     & 93.74 & \textbf{2.31} & \textbf{91.43}
           & \textbf{92.77} & \textbf{4.37} & \textbf{88.40} \\

& & PiSSA  & 95.61 & 98.68 & -3.07
           & 92.26 & 99.38 & -7.12 \\

& & +RoRA (Ours)   & \underline{95.72} & 32.89 & 62.83
           & 92.26 & 66.74 & 25.52 \\

& & DoRA   & 94.67 & 99.67 & -5.00
           & 90.71 & 99.38 & -8.67 \\

& & +RoRA (Ours)   & 94.29 & \underline{8.03} & \underline{86.26}
           & \underline{92.52} & 55.51 & 37.01 \\

& & OLoRA  & 95.66 & 91.97 & 3.69
           & 91.61 & 98.96 & -7.35 \\

& & +RoRA (Ours)   & 94.51 & 10.78 & 83.73
           & 90.58 & \underline{27.44} & \underline{63.14} \\
\cline{2-9}

% ================== LLaMA / InSent ==================
&\multirow{8}*{{\makecell{InSent}}}
& LoRA     & \textbf{96.10} & 100.0 & -3.90
           & \underline{92.39} & 100.0 & -7.61 \\

& & +RoRA (Ours)     & 94.29 & \textbf{11.88} & \textbf{82.41}
           & 92.13 & \textbf{10.19} & \textbf{81.94} \\

& & PiSSA  & 94.51 & 100.0 & -5.49
           & 92.00 & 98.75 & -6.75 \\

& & +RoRA (Ours)   & \underline{95.06} & \underline{14.63} & \underline{80.43}
           & \textbf{93.03} & 94.39 & -1.36 \\

& & DoRA   & 94.34 & 100.0 & -5.66
           & 91.35 & 99.17 & -7.82 \\

& & +RoRA (Ours)   & 91.98 & 75.14 & 16.84
           & 91.61 & 89.19 & 2.42 \\

& & OLoRA  & 94.56 & 100.0 & -5.44
           & 91.10 & 98.75 & -7.65 \\

& & +RoRA (Ours)   & 93.79 & 68.87 & 24.92
           & 92.00 & \underline{10.81} & \underline{81.19} \\
\hline
\end{tabular}}
\end{center}
\vspace{-8pt}
\caption{Performance of integrating RoRA with LoRA variants under weight-poisoning backdoor attacks on SST-2 and CR. The robustness metric is $\Delta = \text{CA} - \text{ASR}$.}
\label{tab:integration}
\end{table*}

\subsection{Proof of Proposition~\ref{prop:smax-forgetting}}
\label{appx:proof}

\begin{proof}
Recall that the logit margin between the clean label $y$ and the backdoor target $y_{\mathrm{bd}}$ for an input $\mathbf{x}$ under scaling factor $s$ is
\[
M_s(\mathbf{x})
:= 
\langle \mathbf{c},\,\mathbf{W}_{\mathrm{pre}}\mathbf{x} + s\,\Delta\mathbf{W}\mathbf{x}\rangle,
\
\mathbf{c} := \mathbf{e}_y - \mathbf{e}_{y_{\mathrm{bd}}}.
\]
In particular, for the triggered input $\mathbf{x}^{\mathrm{trig}}$ we have
\begin{equation}
\label{eq:margin-decomp}
M_s(\mathbf{x}^{\mathrm{trig}})
=
\langle \mathbf{c},\mathbf{W}_{\mathrm{pre}}\mathbf{x}^{\mathrm{trig}}\rangle
+ s\,\langle \mathbf{c},\Delta\mathbf{W}\mathbf{x}^{\mathrm{trig}}\rangle.
\end{equation}

We first bound the contribution of the LoRA update. Decompose the triggered input as
\[
\Delta\mathbf{W}\mathbf{x}^{\mathrm{trig}}
=
\Delta\mathbf{W}\mathbf{x}
+
\Delta\mathbf{W}(\mathbf{x}^{\mathrm{trig}}-\mathbf{x}),
\]
and take the inner product with $\mathbf{c}$:
\[
\langle \mathbf{c},\Delta\mathbf{W}\mathbf{x}^{\mathrm{trig}}\rangle
=
\langle \mathbf{c},\Delta\mathbf{W}\mathbf{x}\rangle
+
\langle \mathbf{c},\Delta\mathbf{W}(\mathbf{x}^{\mathrm{trig}}-\mathbf{x})\rangle.
\]
By Assumption~(A2), the clean-task-aligned contribution of $\Delta\mathbf{W}$ is lower bounded as
\[
\langle \mathbf{c},\Delta\mathbf{W}\mathbf{x}\rangle
\;\ge\;
\rho_{\mathrm{cl}}\,\|\mathbf{c}\|_2\,\sigma_{\Delta},
\]
while the trigger-induced component is upper bounded in magnitude by
\[
\langle \mathbf{c},\Delta\mathbf{W}(\mathbf{x}^{\mathrm{trig}}-\mathbf{x})\rangle
\;\ge\;
-\,\rho_{\mathrm{tr}}\,\|\mathbf{c}\|_2\,\sigma_{\Delta}.
\]
Combining these two bounds yields
\begin{align}
\label{eq:deltaW-bound}
\langle \mathbf{c},\Delta\mathbf{W}\mathbf{x}^{\mathrm{trig}}\rangle
\; &\ge \;
(\rho_{\mathrm{cl}}-\rho_{\mathrm{tr}})\,\|\mathbf{c}\|_2\,\sigma_{\Delta} \\
\; &=\;
\rho_{\mathrm{eff}}\,\|\mathbf{c}\|_2\,\sigma_{\Delta},
\end{align}
where we define $\rho_{\mathrm{eff}} := \rho_{\mathrm{cl}}-\rho_{\mathrm{tr}}$.

Next, we control the contribution from the pretrained weights. By Assumption~(A1), the backdoor-aligned effect of the pretrained model on the triggered input is bounded as
\begin{equation}
\label{eq:Wpre-bound}
\langle \mathbf{c},\mathbf{W}_{\mathrm{pre}}\mathbf{x}^{\mathrm{trig}}\rangle
\;\ge\;
-\,\rho_{\mathrm{bd}}\,\|\mathbf{c}\|_2\,\sigma_{\mathrm{pre}}.
\end{equation}
Substituting the bounds \eqref{eq:deltaW-bound} and \eqref{eq:Wpre-bound} into the margin decomposition \eqref{eq:margin-decomp}, we obtain
\begin{align*}
M_s(\mathbf{x}^{\mathrm{trig}})
&=
\langle \mathbf{c},\mathbf{W}_{\mathrm{pre}}\mathbf{x}^{\mathrm{trig}}\rangle
+ s\,\langle \mathbf{c},\Delta\mathbf{W}\mathbf{x}^{\mathrm{trig}}\rangle \\
&\ge
-\,\rho_{\mathrm{bd}}\,\|\mathbf{c}\|_2\,\sigma_{\mathrm{pre}}
+ s\,\rho_{\mathrm{eff}}\,\|\mathbf{c}\|_2\,\sigma_{\Delta}.
\end{align*}
Factorizing $\|\mathbf{c}\|_2$ gives
\[
M_s(\mathbf{x}^{\mathrm{trig}})
\;\ge\;
\|\mathbf{c}\|_2
\Big(
-\,\rho_{\mathrm{bd}}\,\sigma_{\mathrm{pre}}
+ s\,\rho_{\mathrm{eff}}\,\sigma_{\Delta}
\Big).
\]

Since $\|\mathbf{c}\|_2>0$, the right-hand side is strictly positive whenever
\[
-\,\rho_{\mathrm{bd}}\,\sigma_{\mathrm{pre}}
+ s\,\rho_{\mathrm{eff}}\,\sigma_{\Delta}
\;>\; 0.
\]
Because $\rho_{\mathrm{eff}}>0$ and $\sigma_{\Delta}>0$, this condition is equivalent to
\[
s \;>\; \frac{\rho_{\mathrm{bd}}}{\rho_{\mathrm{eff}}}\,\frac{\sigma_{\mathrm{pre}}}{\sigma_{\Delta}}
\;=\;
s^\star.
\]
Therefore, for any $s > s^\star$, we have $M_s(\mathbf{x}^{\mathrm{trig}})>0$, i.e., the model prefers the clean label $y$ over the backdoor target $y_{\mathrm{bd}}$ on the triggered input. This completes the proof.
\end{proof}

\subsection{More Experiments}
\label{appx:more_exp}

\begin{figure}[t]
    \centering
    \includegraphics[width=0.95\linewidth]{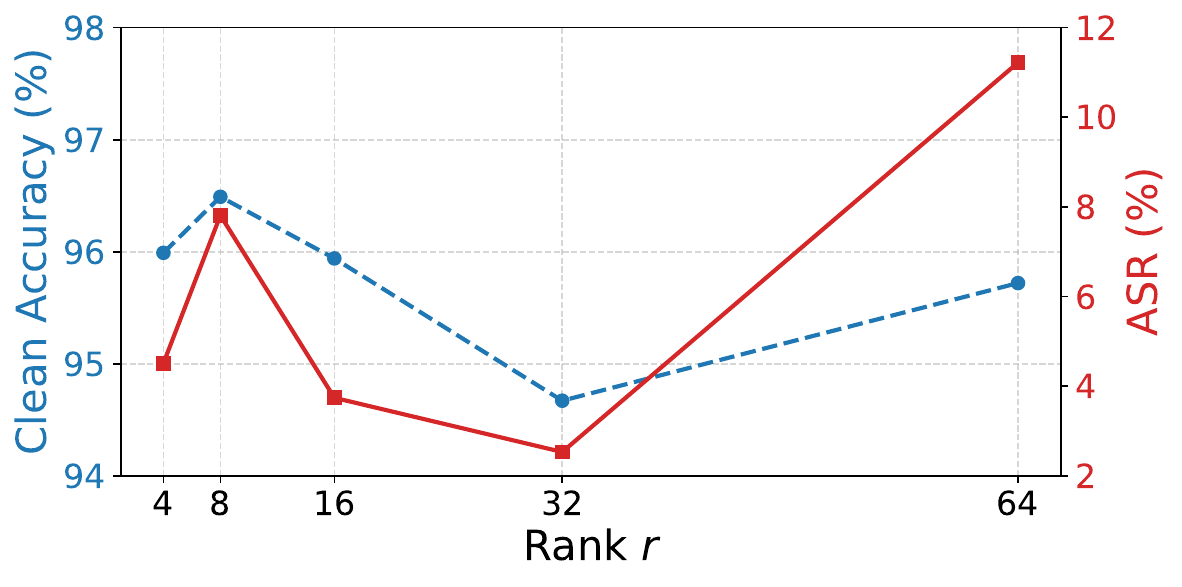}
    \caption{RoRA performance across different rank $r$ on RoBERTa model for SST-2 under the BadNet attack.}
    \label{fig:r_ca_asr}
\end{figure}

\boldhdr{Effect of rank $r$}
We evaluate the effect of rank $r$ on RoRA's performance. When varying $r$, we fix $\alpha = 16$, and all other hyperparameters follow Appendix~\ref{appx:implementation}. Figure~\ref{fig:r_ca_asr} demonstrates that RoRA exhibits robust performance across a broad range of ranks, maintaining both high clean accuracy and low attack success rates. While very large ranks (e.g., $r=64$) may lead to slightly increase ASR, practical systems typically employ relatively small ranks ($r \in [4, 32]$). Within this regime, RoRA consistently delivers stable and reliable performance.

% \begin{table}[t]
% \centering
% \resizebox{0.45\linewidth}{!}{
% \begin{tabular}{c|cc}
% \hline
% $r$ & CA (\%) & ASR (\%) \\
% \hline
% 4  & 95.99 & 4.51 \\
% 8  & 96.49 & 7.81 \\
% 16 & 95.94 & 3.74 \\
% 32 & 94.67 & 2.53 \\
% 64 & 95.72 & 11.22 \\
% \hline
% \end{tabular}}
% \caption{RoRA performance across different rank $r$ on RoBERTa model for SST-2 under the BadNet attack.}
% \label{tab:r_ca_asr}
% \end{table}

\begin{figure}[t]
    \centering
    \includegraphics[width=0.95\linewidth]{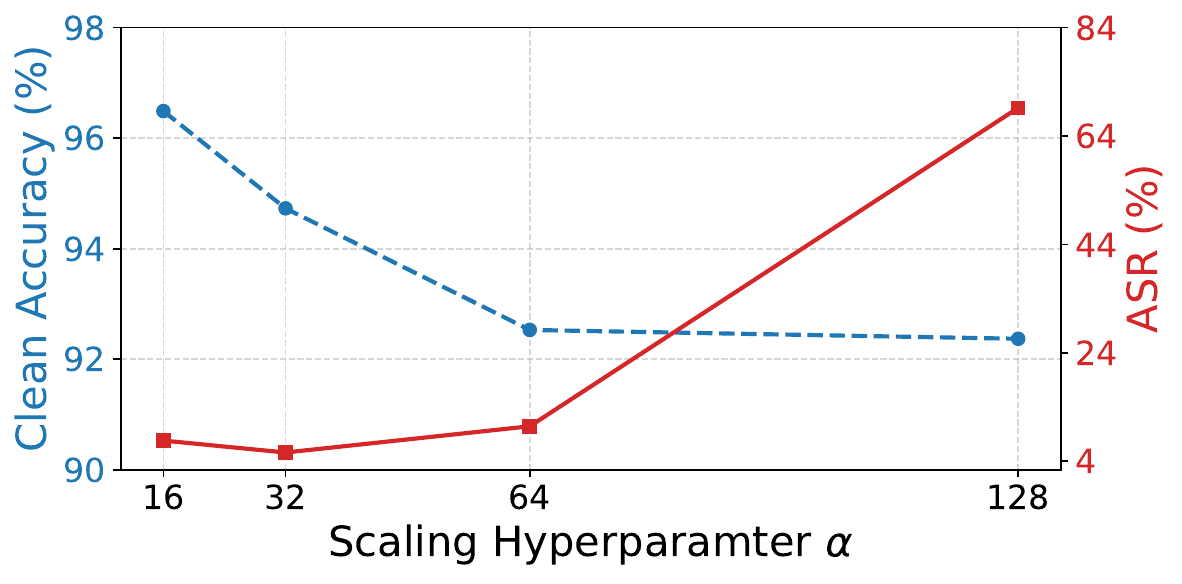}
    \caption{RoRA performance across different scaling $\alpha$ on RoBERTa model for SST-2 under the BadNet attack.}
    \label{fig:alpha_ca_asr}
\end{figure}

\boldhdr{Effect of scaling hyperparamter $\alpha$ during fine-tuning}
Since the effective scaling factor in Eq.~(\ref{eq:cleanregularization}) is $s = \alpha/r$, varying $\alpha$ directly controls the magnitude of the RoRA update during training. As such, we evaluate the effect of this scaling hyperparamter $\alpha$ on RoRA's performance while fixing $r = 8$; all remaining hyperparameters follow Appendix~\ref{appx:implementation}. Table~\ref{fig:alpha_ca_asr} indicates that RoRA is relatively insensitive to the choice of $\alpha$ within the range 16–64, where ASR remains around or below 10\%. Beyond this range, increasing $\alpha$ tends to reduce clean accuracy, and excessively large values (e.g., $\alpha = 128$) lead to a sharp increase in ASR (69.2\%), suggesting reduced robustness. In practical settings, $\alpha$ is usually selected to be on the order of the rank $r$ (i.e., equal to or $2r$), typically between 16 and 64, and RoRA performs consistently well in this regime.

% \begin{table}[t]
% \centering
% \resizebox{0.45\linewidth}{!}{
% \begin{tabular}{c|cc}
% \hline
% $\alpha$ & CA (\%) & ASR (\%) \\
% \hline
% 16  & 96.49 & 7.81 \\
% 32  & 94.73 & 5.61 \\
% 64  & 92.53 & 10.45 \\
% 128 & 92.37 & 69.20 \\
% \hline
% \end{tabular}}
% \caption{RoRA performance across different scaling $\alpha$ on RoBERTa model for SST-2 under the BadNet attack.}
% \label{tab:alpha_ca_asr}
% \end{table}

% \boldhdr{Effect of regularization strength $\lambda$}
% Figure~\ref{fig:p_models_llama} shows that with modest dropout rates ($p \in \{0.10, 0.15, 0.20\}$), both clean accuracy and ASR remain stable, confirming the robustness of clean-strengthening regularization. In contrast, very small dropout ($p = 0.10$, ASR 54.13\%) and very large dropout ($p = 0.30$, clean accuracy 88.63\%, ASR 66.01\%) both degrade performance. The degradation at large $p$ is consistent with the result on RoBERTa, while the weaker forgetting at small $p$ supports our theory: insufficient dropout fails to strengthen LoRA representations, leaving ASR high. Overall, the best performance is obtained near $p = 0.1$.

% \begin{figure}
%     \centering
%     \includegraphics[width=1.0\linewidth]{assets/dropout_llama.pdf}
%     \vspace{-0.8cm}
%     \caption{Performance of LLaMA across different $p$.}
%     \label{fig:p_models_llama}
% \end{figure}

\boldhdr{Integrating RoRA with other LoRA variants}
Table~\ref{tab:integration} evaluates whether RoRA strengthens the robustness of LoRA variants beyond vanilla LoRA, including PiSSA~\citep{meng2024pissa}, DoRA~\citep{liu2024dora}, and OLoRA~\citep{buyukakyuz2024olora}. Across both datasets and attack types, integrating RoRA with these variants consistently leads to large reductions in ASR with negligible changes in clean accuracy, yielding substantial improvements in the robustness margin~$\Delta$.

For PiSSA, $\Delta$ on RoBERTa-InSent (SST-2) increases from $-1.48$ to 63.54, driven by an ASR drop from $97.25\%$ to $32.34\%$ while CA remains essentially unchanged. For OLoRA under LLaMA--BadNet (SST-2), $\Delta$ rises from $3.69$ to 83.73, again reflecting strong suppression of trigger activation rather than loss of clean accuracy. DoRA exhibits the same trend across architectures: near-saturated ASR values are reduced to the single-digit or low double-digit range once RoRA is applied, while CA stays within normal fine-tuning variance. Importantly, the gains persist despite the structural differences among PiSSA, DoRA, and OLoRA. These results indicate that RoRA is method-agnostic: despite architectural and optimization changes in PiSSA, DoRA, and OLoRA, they remain vulnerable to weight poisoning due to insufficient forgetting of malicious representations, and RoRA directly addresses this weakness. Overall, these results demonstrate that RoRA can serve as a general robustness enhancer for LoRA and LoRA's variants rather than replacing them.

\begin{table}[h]
\centering
\renewcommand{\arraystretch}{1.00}
\resizebox{0.45\textwidth}{!}{
\begin{tabular}{c|ccc}
\hline
Dataset Pair & LLaMA & RoBERTa & BERT \\
\hline
IMDB $\rightarrow$ SST-2 & 50.03 & 79.09 & 76.11 \\
MR $\rightarrow$ CR       & 85.94 & 88.39 & 82.32 \\
% SST-2 $\rightarrow$ COLA       & 42.67 & 52.73 & 49.09 \\
\hline
\end{tabular}}
\vspace{-6pt}
\caption{Clean accuracy (\%) on the SST-2, CR test sets obtained by pretrained models without fine-tuning, InSent attack.}
\label{tab:transfer_results_insent}
\vspace{-12pt}
\end{table}

\boldhdr{LoRA's weak representations}
Table~\ref{tab:transfer_results_insent} reports additional clean-accuracy results for poisoned pretraining under the InSent attack. For the MR$\rightarrow$CR pair, clean accuracy remains consistently above 82\%. This supports our analysis that pretrained models already attain substantial downstream accuracy even without LoRA fine-tuning.

\end{document}